\documentclass[11pt]{article}

\usepackage[preprint]{acl}
\usepackage{times}
\usepackage{latexsym}
\usepackage[T1]{fontenc}
\usepackage[utf8]{inputenc}
\usepackage{graphicx}
\usepackage{subcaption}
\usepackage{multirow}
\usepackage{microtype}
\usepackage{inconsolata}

\usepackage{tikz}
\usepackage[dvipsnames]{xcolor}
\usetikzlibrary{calc}
\usetikzlibrary{shapes.geometric}
\usepackage{natbib}
\usepackage[utf8]{inputenc}
\usepackage{wrapfig}
\usepackage{amsmath}
\newtheorem{definition}{Definition}

\usepackage[most]{tcolorbox}

\usepackage{hyperref}       
\usepackage{url}            
\usepackage{booktabs}       
\usepackage{amsfonts}       
\usepackage{nicefrac}       
\usepackage{microtype}      
\usepackage{xcolor}         
\usepackage{paralist}
\usepackage{xspace}
\usepackage{wrapfig}
\usepackage{comment}
\usepackage{pdfcomment}
\usepackage[most]{tcolorbox}
\usepackage{float}
\usepackage{changepage}

\newcommand{\our}{\textsc{ProPer}\xspace}
\newcommand{\xhdr}[1]{\vspace{1mm}\noindent{{\bf #1:\ \ }}}

\newcommand{\gpt}{\textsc{Gpt-5}\xspace}
\newcommand{\gptf}{\textsc{Gpt-4}\xspace}
\newcommand{\llama}{\textsc{LlaMA-8B}\xspace}
\newcommand{\qwen}{\textsc{Qwen-8B}\xspace}
\newcommand{\papertitle}{\our Agents: \underline{Pro}activity Driven \underline{Per}sonalized Agents for Advancing Knowledge Gap Navigation}

\title{\papertitle}
\author{
Kirandeep Kaur$^{1,*}$ \quad
Vinayak Gupta$^{1,*}$\footnotemark[2] \quad
Aditya Gupta$^{2}$ \quad
Chirag Shah$^{1}$ \\
$^{1}$University of Washington, Seattle, WA, USA \\
$^{2}$Issaquah High School, Issaquah, WA, USA \\
Corresponding Author: \texttt{kaur13@cs.washington.edu} \\
}

\begin{document}
\maketitle

\begingroup
\renewcommand{\thefootnote}{\fnsymbol{footnote}}
\footnotetext[1]{Equal contribution.}
\footnotetext[2]{Current affiliation: Chan Zuckerberg Biohub.}
\endgroup

\begin{abstract}
Current approaches to proactive assistance move beyond the \textit{ask-and-respond} paradigm by anticipating user needs. In practice, they either burden users with clarifying questions or rely on context-based extrapolation, often leading to unnecessary or mistimed interventions. Such systems lack explicit mechanisms to model users' knowledge gaps, resulting in incomplete or suboptimal task outcomes. To address this, we propose \textbf{\our}, a framework that explicitly models user-specific knowledge gaps in a controlled manner. Central to our approach is the notion of \textit{dimensions}: structured, task-relevant factors that define the considerations required for effective task completion. Given a user query, the \textbf{DGA} (Dimension Generating Agent) identifies \textit{explicit dimensions} (from the user's query) and generates a set of candidate \textit{implicit dimensions} capturing unarticulated aspects of the task. The \textbf{RGA}  (Response Generating Agent) integrates both explicit and implicit dimensions selectively to produce personalized, context-aware, and proactively informative responses. We evaluate \our across multiple domains using a structured, gap-aware rubric that measures coverage, initiative appropriateness, and intent alignment.  \our improves on quality scores and win rates across all domains, achieving up to 84\% gains in single-turn evaluation and consistent dominance in multi-turn interactions. All code for \our\ is available at: \url{https://github.com/i-kiran/ProPer-Agent}.
\end{abstract}

\section{Introduction}

\begin{figure*}[t!]
    \centering
    \includegraphics[width=0.99999\linewidth]{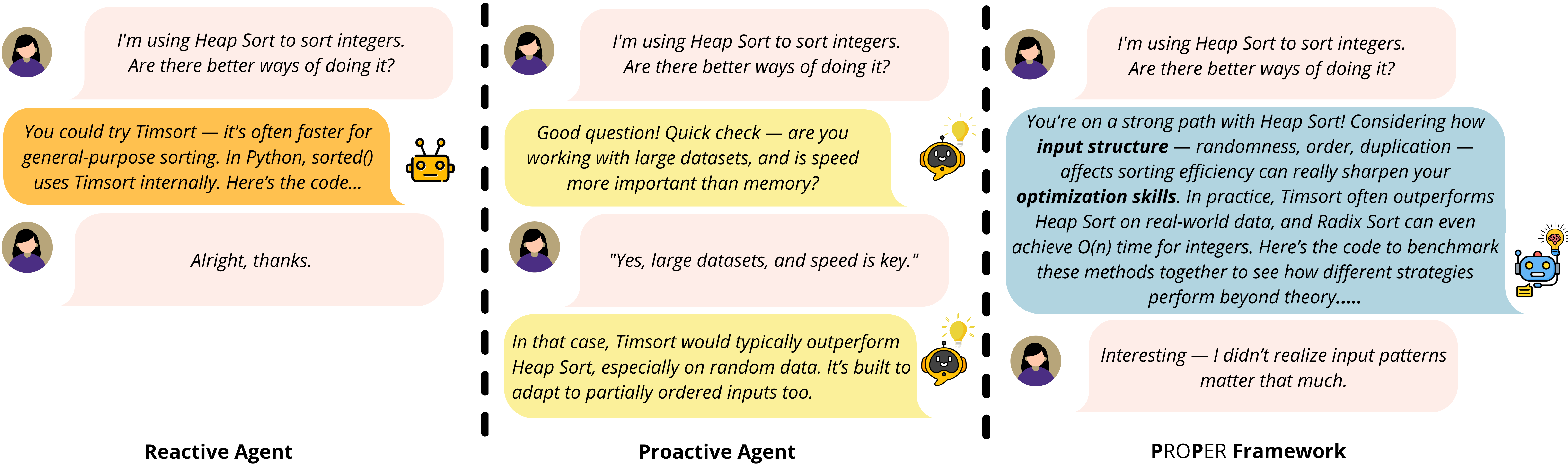}
    \caption{Different agent responses to a user query. The \textbf{Reactive Agent} provides immediate task fulfillment without exploring user context, goals, or learning needs. The \textbf{Proactive Agent} clarifies task-related ambiguities to optimize the immediate solution but remains confined to the user's knowledge space. \our, on the other hand, embodies a higher-order interaction strategy established through a learning-centric response structure.} 
    \label{fig:motivation}
    \label{fig:motivation}
\end{figure*}
While traditional conversational agents rely on end users to express their needs, proactive agents move beyond \textit{reactive} interactions by anticipating user needs. In practice, most approaches remain anchored in what users explicitly express or can readily surface. Clarification-based interventions address either \emph{known unknowns}, by eliciting missing specifications \citep{zhang-etal-2024-ask,deng-etal-2024-dont,hahn2025proactive}, or \emph{unknown knowns}, by prompting users to surface implicit assumptions \citep{Huang2024PromptingEA}. Other approaches extrapolate from observable context or environment \citep{lu2024proactiveagent,dong-etal-2025-protod}. However, such proactive interventions risk being mistimed or misaligned, degrading trust and interaction quality \citep{zargham2022understanding,Chen2025NeedHelp,diebel2025ai,Pu2025Assistance}.

This limitation stems from the absence of a deeper understanding of users’ knowledge and ignorance. \textit{Knowledge gaps}, task-relevant considerations that lie outside the user’s awareness and are not readily inferable from context, are critical for successful task completion. Failing to account for such gaps can lead to incomplete or suboptimal outcomes. We refer to such gaps as \emph{unknown unknowns}~\cite{none_too_solid}, i.e., unknowns with respect to a user. 

Furthermore, what a user knows determines what they do not know~\cite{kaur2026knowing}. We operationalize this by identifying \textit{dimensions}: task-related aspects that represent important considerations relevant to a given query. To this end, \textbf{DGA}, a fine-tuned LLM, learns to model the distribution of dimensions across domain-specific tasks. Given a user’s history and query, it extracts explicit dimensions from the input and then identifies implicit dimensions, i.e., user knowledge gaps, using the learned distribution. Figure~\ref{fig:motivation} illustrates this process. For a simple sorting query, beyond the explicitly stated algorithm choice, such dimensions include input structure (e.g., randomness, duplicates), performance constraints, and scalability considerations. 

While such gaps are often critical, introducing them indiscriminately risks disrupting user intent. We therefore frame effective assistance as a problem of \emph{calibrating personalization and proactivity}. \our comprises an \textbf{activation module} that evaluates the candidate dimensions against their relevance, timeliness, and compatibility with the user’s expressed intent. The activated dimensions are then provided to \textbf{RGA}, which conditions generation on this curated set to ensure that additional information is introduced in a focused and non-disruptive manner. 

Finally, we evaluate \our using a structured, gap-aware rubric designed to assess proactive assistance beyond surface correctness. The rubric measures response quality along three dimensions: coverage of task-relevant knowledge gaps, appropriateness of initiative, and alignment with user intent. Our results show that \our consistently improves task utility over strong base LLMs and CoT prompting, especially in medical tasks where hidden risks, constraints, and user needs matter. The gains are smaller in tightly defined tasks like coding. To summarize, the key contributions are:
\begin{compactitem}
    \item We formalize \emph{proactivity as a calibration problem}, emphasizing selective intervention over users' knowledge gaps.
    \item We introduce a \emph{dimension-based representation} that captures task-relevant considerations beyond the user’s expressed query, enabling explicit modeling of knowledge gaps.
    \item We propose \our, a modular agent architecture that decouples knowledge gap discovery from response generation to balance personalization and proactivity.
    \item  We develop a gap-aware evaluation framework and demonstrate consistent improvements across across medical, recommendation, and coding domains.
\end{compactitem}

\section{Related Work}
\vspace{-0.41em}
Proactive conversational agents extend beyond explicit user queries by initiating assistance. Clarification-based dialogue systems refine underspecified requests by eliciting additional input, addressing \emph{known unknowns} \citep{zhang-etal-2024-ask,deng-etal-2024-dont,hahn2025proactive, kaur2026knowing}. Structured prompting and planning-based methods further improve reasoning through internal deliberation or decomposition, but remain limited to information that is explicitly stated or readily inferable \citep{deng-etal-2023-prompting,Liu2024ProactiveCA,madaan2023self}. Recent approaches extend proactivity through autonomous planning, task monitoring, or adaptive policies \citep{dong-etal-2025-protod,lu2024proactiveagent, li2025knowledge}, including multimodal and embodied agents that leverage environmental signals \citep{Bandyopadhyay2025YETIT,hahn2025proactive}. While these systems move beyond reactive behavior, their proactivity is primarily driven by observable state, heuristics, or task progress, rather than modeling unarticulated user needs \citep{10.1609/aaai.v38i16.29710,park2024generativeagents,ribeiro2020beyondaccuracy, jiang2024unknown, griot2025metacognition}. 

Human-centered studies emphasize that poorly calibrated proactivity can undermine user trust, agency, and interaction quality when assistance is mistimed or misaligned \citep{zargham2022understanding,Chen2025NeedHelp,diebel2025ai,Pu2025Assistance,10.1145/3626772.3657843}. In parallel, personalization and user modeling~\cite{park2024generativeagents,ribeiro2020beyondaccuracy}capture latent preferences or goals from history, but remain constrained to user's problem space rather than augmenting their knowledge itself.  As a result, existing approaches do not actively model or address unarticulated knowledge gaps in user interactions. 

On the other hand, aligning proactive steps to achieve a goal has been done in several other domains, such as text-to-image generation~\cite{hahn2025proactive}, time-series modeling~\cite{cheng2025atsf, guptaproactive, gupta_proper_tist}, dialogue clarification~\cite{deng-etal-2023-prompting}, recommender systems~\cite{proactive_recs, kaur2026efficient, gupta2022doing}, and goal-decomposition in multi-agent planning~\cite{li2025agentoriented}.
\section{Problem Formulation}
\label{sec:probformulation}
Let $q \in \mathcal{Q}$ denote the current user query, $h \in \mathcal{H}$ the interaction history, and $p \in \mathcal{P}$ persona-related explicit attributes. We define the user state as $u = (q, h, p) \in \mathcal{U}$. Given $u$, a baseline system produces an initial response $r_0 \in \mathcal{R}$ based on the explicitly available information. However, some task-relevant aspects may remain unarticulated in $(q,h)$ or unaddressed in $r_0$. The objective is to generate a final response $r \in \mathcal{R}$ by selectively intervening when warranted by the user state.

\begin{definition}[Interaction Dimensions]
\label{def:interaction_dimensions}
Let $\mathcal{D}$ denote a shared universe of interaction dimensions.  
Given an interaction state $I = (u, r_0)$, the set of interaction dimensions is defined as:
\begin{equation}
\mathcal{D}(I)
:=
\mathcal{D}^{\text{user}}_{\text{exp}}(u)
\;\cup\;
\mathcal{D}^{\text{sys}}_{\text{exp}}(r_0)
\;\cup\;
\mathcal{D}_{\text{imp}}(u),
\end{equation}
where,
\begin{align}
\mathcal{D}^{\text{user}}_{\text{exp}}(u)
&=
\{ d \in \mathcal{D} \mid d \prec (q,h) \}, \\
\mathcal{D}^{\text{sys}}_{\text{exp}}(r_0)
&=
\{ d \in \mathcal{D} \mid d \prec r_0 \}, \\
\mathcal{D}_{\text{imp}}(u)
&=
\{ \tilde d \in \mathcal{D}
\mid
\tilde d \not\prec (q,h)
\;\land\;
\tilde d \sim u
\}.
\end{align}
\end{definition}

Here, $\mathcal{D}(I)$ comprises user-explicit dimensions $\mathcal{D}^{\text{user}}_{\text{exp}}(u)$, system-explicit dimensions $\mathcal{D}^{\text{sys}}_{\text{exp}}(r_0)$, and implicit task-relevant dimensions $\mathcal{D}_{\text{imp}}(u)$, enabling reasoning about both unaddressed explicit needs and unarticulated knowledge gaps.

\begin{definition}[Selective Activation]
\label{def:selective_activation}
Given an interaction state $I=(u,r_0)$, selective activation identifies a set of interaction dimensions $\mathcal{D}_{\text{act}}(I) \subseteq \mathcal{D}(I)$ such that
\begin{equation}
\mathcal{D}_{\text{act}}(I)
\subseteq
\Big(
\mathcal{D}^{\text{user}}_{\text{exp}}(u)
\setminus
\mathcal{D}^{\text{sys}}_{\text{exp}}(r_0)
\Big)
\;\cup\;
\mathcal{D}_{\text{imp}}(u).
\end{equation}
\end{definition}
The set $\mathcal{D}_{\text{act}}(I)$ captures dimensions that warrant proactive intervention, including user-explicit dimensions not addressed by the baseline response and task-relevant implicit dimensions that remain unarticulated. Selective activation thus unifies two sources of incompleteness: unmet explicit needs and latent knowledge gaps.
\begin{definition}[Post-hoc Calibrated Ranking]
\label{def:calibrated_ranking}
Given an interaction state $I=(u,r_0)$ and a selectively activated candidate set $\mathcal{D}_{\text{act}}(I)$, post-hoc calibrated ranking defines the problem of selecting a budgeted subset of interaction dimensions $S_k \subseteq \mathcal{D}_{\text{act}}(I)$, where $k$ denotes a fixed intervention budget.
\end{definition}

The selected set $S_k$ represents the dimensions that warrant proactive intervention in the final response. Calibrated ranking balances three considerations: quality, whether it addresses explicit user needs, and the degree of redundancy.
\begin{definition}[\our Calibration]
\label{def:proper}
Given $I=(u,r_0)$, user's interaction state and
$
\mathcal{D}(I)
=
\mathcal{D}^{\text{user}}_{\text{exp}}(u)
\;\cup\;
\mathcal{D}^{\text{sys}}_{\text{exp}}(r_0)
\;\cup\;
\mathcal{D}_{\text{imp}}(u)
$, the corresponding interaction dimensions.
Let
$
U_u := \mathcal{D}^{\text{user}}_{\text{exp}}(u)
\quad\text{and}\quad
S_k^{*}(I) \subseteq \mathcal{D}_{\text{act}}(I)
$
denote the user-explicit dimensions and a calibrated set of activated dimensions, respectively.
Let $
\mathcal{F}:
\mathcal{R}
\times
\mathcal{U}
\times
\mathcal{R}
\times
2^{\mathcal{D}}
\times
2^{\mathcal{D}}
\rightarrow
\mathbb{R}
$
be a utility function.
The learning objective is
\begin{equation}
\label{eq:proper_objective}
\hat r
=
\arg\max_{r\in\mathcal{R}}
\mathcal{F}\!\left(r \mid u,\, r_0,\, U_u,\, S_k^{*}(I)\right).
\end{equation}
\end{definition}
We formulate proactive assistance as a calibration problem over response repair, asking when and which unaddressed or implicit dimensions should be introduced given a user state and baseline response. The objective $J(\cdot)$ defines this selection problem, and the utility $\mathcal{F}$ evaluates candidate responses conditioned on the selected dimensions.

\section{Proposed Methodology}
\begin{figure*}
    \centering
    \includegraphics[width=0.95\linewidth]{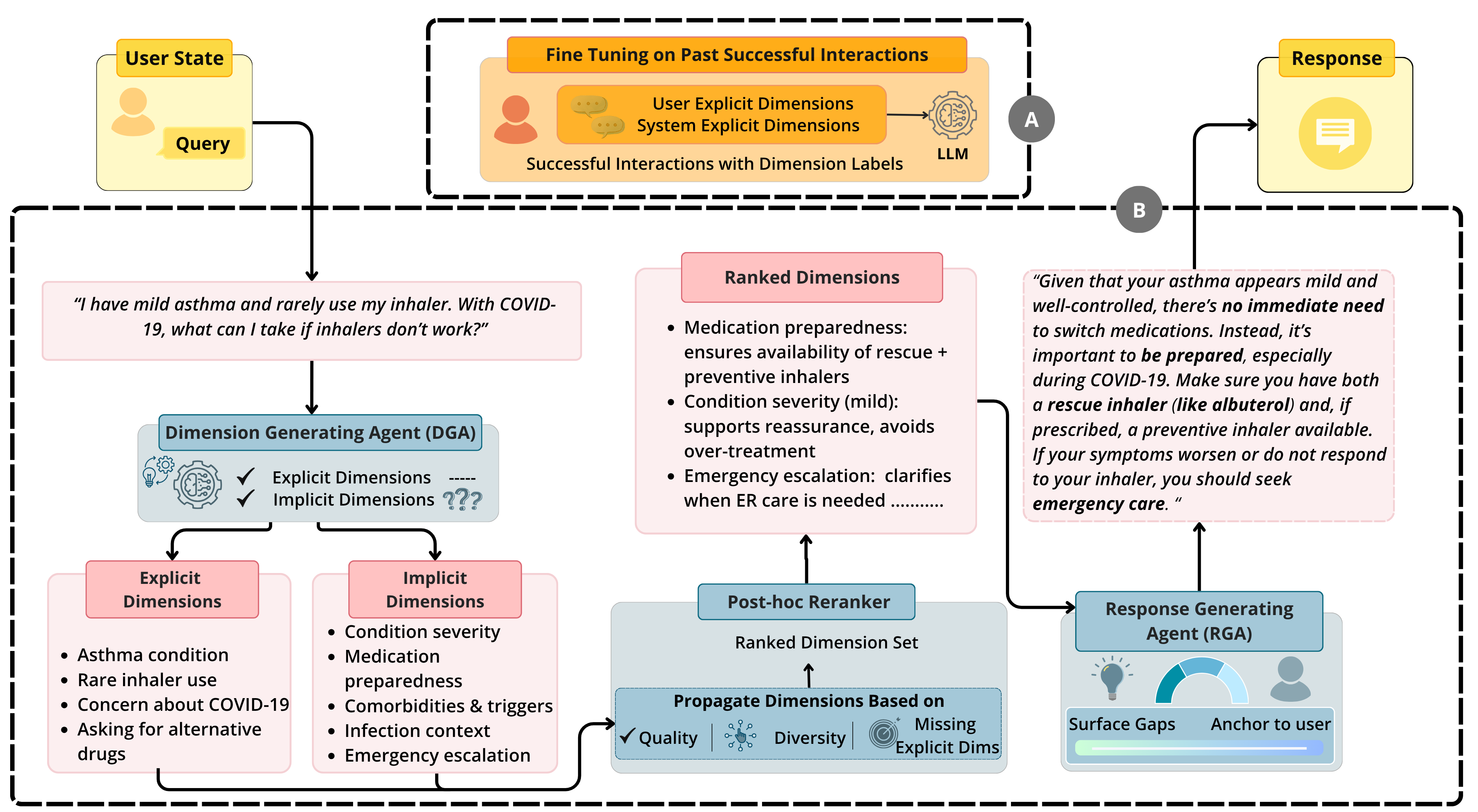}
    \caption{Overview of the \our framework. During \textbf{training} (A), the \textbf{Dimension Generating Agent (DGA)} is fine-tuned on successful interactions annotated with \textbf{user-explicit} and \textbf{system-explicit dimensions}, learning task-specific priors. At \textbf{inference} (B), the DGA identifies \textbf{explicit} and \textbf{implicit candidate dimensions} from the user state. A \textbf{post-hoc reranker} selects a calibrated subset, and the \textbf{Response Generating Agent (RGA)} updates the base response by selectively integrating them, balancing \textbf{proactivity} with \textbf{user intent}.}
    \label{fig:placeholder}
\end{figure*}

This section describes our methodology, which operationalizes the formulation in Section~\ref{sec:probformulation} via three stages.
First, we construct \emph{interaction dimensions} by separating user-explicit dimensions, system-covered dimensions, and implicit task-relevant gaps (Def.~\ref{def:interaction_dimensions}).
Second, we \emph{selectively activate} a candidate set and perform \emph{calibrated ranking} to obtain a budgeted intervention set (Def.~\ref{def:selective_activation} and Def.~\ref{def:calibrated_ranking}).
Finally, we generate a response conditioned on the user-explicit dimensions and the calibrated intervention set (Def.~\ref{def:proper}).

\subsection{Dimension Generating Agent (DGA)}
\label{sec:method_dga}
DGA is responsible for inferring implicit dimensions $\mathcal{D}_{\text{imp}}(u) \subseteq \mathcal{D}$ that correspond to task-relevant knowledge gaps. For this, we fine-tune an LLM using supervision derived from successful assistance trajectories, where interactions are annotated with dimension-level labels (details in Appendix~\ref{app:data_pipeline}). Through this, DGA learns to generate plausible missing dimensions conditioned on the user state.

At inference time, given $u=(q,h,p)$, the DGA produces a set of candidate implicit dimensions $\mathcal{D}_{\text{imp}}(u)$ along with log probabilities as confidence scores (see Appendix~\ref{app:dimension_schema} for the exact prompt used for DGA). User-explicit dimensions $\mathcal{D}^{\text{user}}_{\text{exp}}(u)$ and baseline-covered dimensions $\mathcal{D}^{\text{sys}}_{\text{exp}}(r_0)$ are extracted separately, enabling construction of $\mathcal{D}(I)$ for $I=(u,r_0)$ (Def.~\ref{def:interaction_dimensions}).

Through fine-tuning, DGA goes beyond text to internalize all the key dimensions considered to assist a user. At inference time, given a user query, DGA leverages its internalized dimensions of other tasks to produce a set of candidate interaction dimensions $\mathcal{D}(u)$ which were used for successful assistance and the current user might be missing.

\subsection{Post-hoc Calibrated Reranker} \label{sec:method_reranker}
The reranker determines which interaction dimensions inferred by the DGA should be activated for response generation. It operationalizes selective activation and calibrated ranking (Def.~\ref{def:selective_activation}, Def.~\ref{def:calibrated_ranking}), treating proactivity as a controlled decision.

Given an interaction state $I=(u,r_0)$, the reranker constructs the activation pool $\mathcal{D}_{\text{act}}(I)$ comprising (i) user-explicit dimensions not addressed by the baseline response and (ii) diverse and non-redundant implicit task-relevant dimensions. All dimensions are encoded using BGE-small embeddings. From this pool, the reranker selects a budgeted set $S_k^{\ast}(I)$ by maximizing the objective in Eq.~\ref{eq:posthoc_ranker}. The objective favors dimensions with high DGA confidence, encourages alignment with unmet explicit user needs via semantic similarity, and penalizes redundancy among selected dimensions to promote complementary coverage. The budget $k$ controls the degree of proactivity by limiting how many dimensions are activated.

{
\small
\begin{equation}
\label{eq:posthoc_ranker}
\begin{aligned}
S_k^{\ast}(I)
&=
\arg\max_{S \subseteq \mathcal{D}_{\text{act}}(I),\, |S|=k}
\Bigg[
\underbrace{\sum_{d \in S} \log P(d \mid u)}_{\text{quality}}
\\
&\,-
\lambda_1
\underbrace{\sum_{d \in S}
\max_{e \in \mathcal{E}(I)} \mathrm{sim}(d,e)}_{\text{unmet explicit alignment}}
 -
\lambda_2
\underbrace{\sum_{i<j} \mathrm{sim}(d_i,d_j)}_{\text{implicit diversity}}
\Bigg].
\end{aligned}
\end{equation}
}
\noindent
where $I=(u,r_0)$ and
$\mathcal{E}(I)=\mathcal{D}^{\text{user}}_{\text{exp}}(u)\setminus
\mathcal{D}^{\text{sys}}_{\text{exp}}(r_0)$.
All similarity terms use cosine similarity over BGE-encoded dimension representations.

\subsection{Response Generating Agent (RGA)}
\label{sec:method_rga}
The Response Generating Agent (RGA) generates the final system response by selectively updating a baseline response using the calibrated set of activated interaction dimensions. It is implemented as a prompt-driven generation module that conditions on the user query, interaction context, the baseline response, and the activated dimensions selected by the post-hoc reranker.

Given an interaction state $I=(u,r_0)$ and a calibrated intervention set $S_k^{\ast}(I)$, the RGA produces an updated response $\hat r$ that preserves the intent and structure of $r_0$ while selectively incorporating additional information. The prompt explicitly encodes \emph{initiative calibration}: the RGA infers how proactive to be from the user query itself, expanding when signals of confusion, risk, or uncertainty are present, and remaining focused when the query is narrow or constrained.

The prompt further distinguishes between explicit and implicit dimensions, requiring explicit gaps to be addressed by default while incorporating implicit gaps only when warranted by the interaction context. To avoid overreach, the RGA is constrained to favor concise additions over full rewrites and to ask at most one clarifying question when resolving an implicit gap would require user-specific information. These constraints ensure that proactivity remains targeted, non-assumptive, and aligned with user intent.

We use domain-specific instantiations of this prompt for coding assistance, clinical support, and recommendation settings. The full prompts for all three domains are provided in Appendix~\ref{app:rga_prompts}.

\subsection{\our Framework}

We now describe the end-to-end execution of the \our framework. The process begins with the construction of a user state composing user query, interaction history, and persona related information (if available). Together, these components define the interaction context used throughout the pipeline. A baseline system response \(r_0\) is generated using a standard assistant, providing a reference point for later phases. The first stage is handled by the Dimension Generating Agent (DGA). Given the interaction context, the DGA infers a set of interaction dimensions \( \mathcal{D}(u) \) by distinguishing between explicit dimensions surfaced in the interaction and implicit gaps, task-relevant dimensions that are not currently addressed. The DGA outputs a pool of candidate dimensions along with confidence scores reflecting their relevance under the user state.

In the second stage, a post-hoc calibrated reranker selects a budgeted subset \( S_k^{\ast}(u,r_0) \) by optimizing quality, alignment with user-explicit needs, and diversity. This frames proactivity as a controlled choice of which missing aspects to address, rather than simple response expansion.

Finally, RGA takes the input user query, the baseline response, explicit dimensions, and the calibrated set of activated dimensions. Conditioned on this structured guidance, the RGA produces an updated response \( \hat r \) that preserves the intent and structure of the baseline output while selectively addressing the activated dimensions. Explicit dimensions anchor the response to articulated user intent, while activated dimensions guide targeted expansion over task-relevant aspects that were previously unaddressed.


\section{Experiments} \label{sec:experiments}
We evaluate \our to examine whether \emph{calibrated activation of implicit, task-relevant dimensions} improves assistance quality beyond reactive and clarification-based baselines. Our evaluation focuses not only on response correctness, but also on whether proactive intervention is timely, appropriate, and aligned with user needs. 

\noindent Our experiments aim to answer the following research questions (RQ):
\begin{compactitem}
    \item[\textbf{RQ1}] Does \our improve end-to-end task utility across domains?
    \item[\textbf{RQ2}] How do \textsc{DGA}, reranking, and \textsc{RGA} individually contribute to performance?
    \item[\textbf{RQ3}] Are observed gains driven by calibrated proactivity rather than increased verbosity?
    \item[\textbf{RQ4}] Does \our remain robust in multi-turn conversational settings?
\end{compactitem}

Unless otherwise specified, \textsc{DGA} and \textsc{RGA} share the same underlying LLM. All experiments were conducted on an NVIDIA H100 GPU.

\begin{table*}[t!]
\small
\centering
\begin{tabular}{lcccccc}
\toprule
\textbf{Dataset $\rightarrow$} & \multicolumn{2}{c}{\textbf{Medical (MD)}} & \multicolumn{2}{c}{\textbf{Code-Contests}} & \multicolumn{2}{c}{\textbf{PWAB}} \\
\cmidrule(lr){2-3} \cmidrule(lr){4-5} \cmidrule(lr){6-7}
\textbf{Models $\downarrow$} & $\mu\text{Score}$ & \text{Win\%} & $\mu\text{Score}$ & \text{Win\%} & $\mu\text{Score}$ & \text{Win\%} \\
\midrule
\llama                           & 2.19 & 10.52 & 1.26 & 15.51 & 2.34 & 6.83    \\
\hspace{2mm}vs \llama + \our     & \textbf{3.86}\textsuperscript{$\dagger$} & \textbf{89.48}\textsuperscript{$\dagger$} & \textbf{2.13}\textsuperscript{$\dagger$} & \textbf{84.49}\textsuperscript{$\dagger$} & \textbf{4.06}\textsuperscript{$\dagger$} & \textbf{93.17}\textsuperscript{$\dagger$}   \\
\midrule
\qwen                           & 2.93 & 18.73 & 2.24 & 24.76 & 3.12 & 12.50    \\
\hspace{2mm}vs \qwen + \our     & \textbf{4.03}\textsuperscript{$\dagger$} & \textbf{81.27}\textsuperscript{$\dagger$} & \textbf{2.84}\textsuperscript{$\dagger$} & \textbf{75.24}\textsuperscript{$\dagger$} & \textbf{4.29}\textsuperscript{$\dagger$} & \textbf{87.50}\textsuperscript{$\dagger$}   \\
\midrule
\gptf                & 3.28 & 29.74 & \textbf{3.19}\textsuperscript{$\dagger$} & \textbf{68.93}\textsuperscript{$\dagger$} & 3.46 & 23.61 \\
\hspace{2mm}vs \llama + \our  & \textbf{3.73} & \textbf{70.26}\textsuperscript{$\dagger$} & 2.08 & 31.07 & \textbf{4.11} & \textbf{76.39}\textsuperscript{$\dagger$} \\
\midrule
\gptf                & 3.26 & 19.26 & \textbf{3.11} & 43.63 & 3.53 & 17.40 \\
\hspace{2mm}vs \qwen + \our     & \textbf{4.03}\textsuperscript{$\dagger$} & \textbf{80.74}\textsuperscript{$\dagger$} &2.97 & \textbf{56.37} & \textbf{4.24}\textsuperscript{$\dagger$} & \textbf{82.60}\textsuperscript{$\dagger$} \\
\bottomrule
\end{tabular}
\caption{One-on-one comparison of \llama, \qwen, \gptf, and our model, with \gpt as judge. $\boldsymbol{\mu}\text{\textbf{Score}}$ denotes the average score across all entries; \textbf{\text{Win\%}} denotes the percentage of samples where the model outperformed the paired model. \our improves over base LLMs and surpasses GPT-4 in Medical (MD) and PWAB; gains are smaller in Code-Contests, as any model can find optimal solutions. \textsuperscript{$\dagger$} indicates $p \le 0.01$.}
\label{tab:pairwise_results}
\end{table*}

\subsection{Experimental Setup}
\xhdr{Datasets}
We evaluate \our across three domains that differ in how and when proactive assistance is appropriate.
\textbf{Medical (MD)}~\cite{medDG} involves patient-facing medical queries related to diagnosis, treatment, or symptoms, where proactive behavior must be cautious, uncertainty-aware, and emotionally appropriate.
\textbf{Code-Contests}~\cite{li2022competition} contains programming tasks in which coding assistance.
\textbf{PWAB}~\cite{cai2024personalwab} is an online Amazon-shopping dataset comprising mulitiple recommendation and web search queries.
Together, these datasets stress complementary aspects of calibrated proactivity, ranging from restraint (medical) to guidance (coding) to preference balancing (shopping).
Additional preprocessing details are provided in Appendix~\ref{app:data_pipeline}.

\xhdr{Baselines}
Following prior work on proactive agents~\cite{lu2024proactiveagent}, we use \llama-3.1-8B-Instruct~\cite{grattafiori2024llama} and Qwen-3.1-8B~\cite{qwen3} as our primary baselines. These models represent strong, instruction-tuned LLMs of comparable scale and are evaluated both as standalone assistants and as the backbone architectures for \our. This design isolates the effect of calibrated dimension generation and response orchestration from raw model capacity. Full fine-tuning configurations and hyperparameters for \textsc{DGA} are provided in Appendix~\ref{app:dga_training}.

\xhdr{Evaluation Protocol}
We evaluate responses using an external LLM-based judge (\gpt{}), following recent work on holistic evaluation of open-ended assistance. For each input, we generate responses from all compared models and ask \gpt{} to assign a quality score on a 0--5 scale, accompanied by a brief justification. The evaluation rubric emphasizes overall helpfulness, relevance, and the appropriateness of proactive guidance, rather than surface-level correctness alone.

We report the mean quality score across samples ($\mu\text{Score}$) and the percentage of samples in which a model achieves the highest score (Win\%). Unless otherwise specified, evaluations consider complete responses rather than isolated facts, reflecting our focus on calibrated assistance.

\begin{figure}[t!]
\centering
\begin{subfigure}{\columnwidth}
  \centering
  \includegraphics[width=\columnwidth]{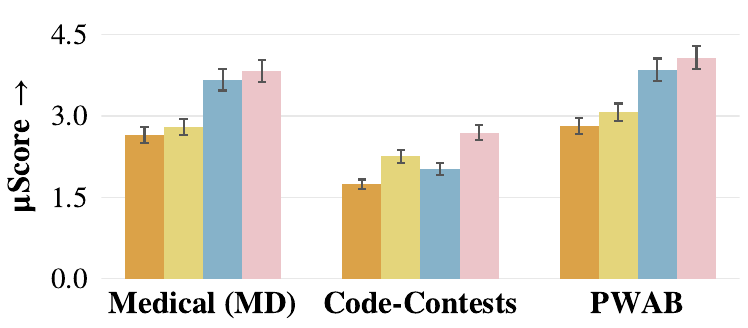}
\end{subfigure}

\vspace{0.5em}
\begin{subfigure}{\columnwidth}
  \centering
  \includegraphics[width=\columnwidth]{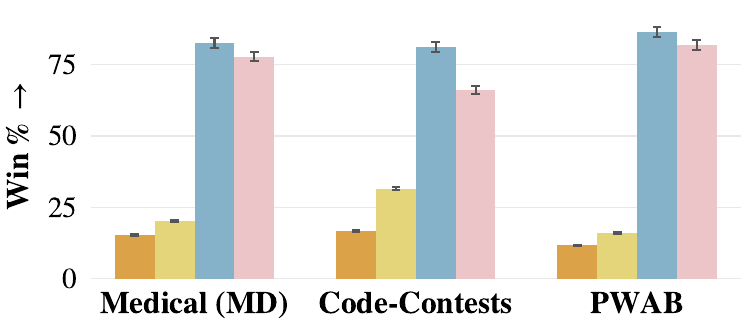}
\end{subfigure}

\centering
\includegraphics[width=0.9\columnwidth]{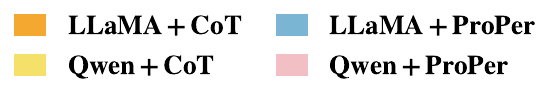}
\vspace{-0.3cm}
\caption{Comparison of \our with CoT prompting applied to \llama and \qwen. \our consistently outperforms other models even when CoT prompting enhances baseline LLMs.}
\label{fig:cot}
\vspace{-0.2cm}
\end{figure}

\subsection{Results and Discussions}
\xhdr{RQ1: End-to-End Task Utility} To evaluate how \our improves proactivity among base LLMs, we compare the responses generated by the base LLM (\llama or \qwen) with those from \our with the same LLM as the base. We also include \gptf for reference. Table~\ref{tab:pairwise_results} results clearly show that, on average, \our outperforms the base LLM on 84\% of entries across all datasets. \our achieves substantially higher mean scores and win rates when paired with either \llama or \qwen, reflecting the importance of proactively surfacing unarticulated risks, constraints, and preferences that are often omitted from the initial query. In contrast, improvements on Code-Contests are more modest: while \our still wins more frequently than base models, gains in mean score are smaller, suggesting diminishing returns from proactive intervention in highly structured tasks with well-defined solutions. 
In addition, we also compare \our with chain-of-thought (CoT) prompting applied to \llama and \qwen (refer to Appendix~\ref{app:cot_prompt} for COT prompts). Following ~\cite{madaan2023self}, we construct a chain of prompting that first asks the LLM to generate a response to a user query, then analyze which aspects the response did not sufficiently address, and finally reiterate the process to improve the response. The comparison of CoT models with \our, shown in Figure~\ref{fig:cot}, show that even after CoT prompting, which doe improve the base LLM, \our still outperforms. In addition, while CoT improves standard fine-tuned models, the gains are marginal. To complement the quantitative results, we examine a representative query from the Medical (MD) dataset and report the full responses generated by different models below.

\begin{tcolorbox}
{
\small
\begin{adjustwidth}{-0.9cm}{-0.9cm}
\begin{quote}\itshape
\noindent\textbf{User Query:} Because I have type 1 diabetes, what is likely to happen to me if I get the coronavirus?
\end{quote}
\begin{center}\rule{0.4\linewidth}{0.8pt}\end{center}

\begin{quote}\itshape
\noindent\textbf{\llama (Fine-Tuned):} People with diabetes are at increased risk for severe illness and death from COVID-19. It is not a guarantee that you would die from it, but the risk is higher. The virus spreads mainly through droplets of saliva, mucus, and other respiratory secretions, typically released when an infected person coughs, sneezes, or breathes.
\end{quote}

\begin{quote}\itshape
\noindent\textbf{\gptf:} If you have type 1 diabetes and get COVID-19, you'll most likely have mild symptoms and recover, especially if you're young. The main issue is that illness can cause higher blood sugars and ketones, so you may need more insulin and closer monitoring. Serious problems are uncommon when diabetes is well managed.
\end{quote}

\begin{quote}\itshape
\noindent\textbf{\llama + \our:} Because you have Type 1 diabetes, COVID-19 might be more severe. Your immune system may struggle raising the risk of breathing complications, and can affect blood sugar control. Vaccination helps, avoid crowds, practice hygiene, and follow your doctor's advice. Other conditions like hypertension, kidney, or heart disease increase risk.
\end{quote}
\end{adjustwidth}
}
\end{tcolorbox}

\begin{table*}[t!]
\small
\centering
\setlength{\tabcolsep}{3pt}
\resizebox{\linewidth}{!}{
\begin{tabular}{c|cc|cc|cc}
\toprule
\multirow{2}{*}{$\boldsymbol{(\lambda_1,\lambda_2)}$}
& \multicolumn{2}{c|}{\textbf{Medical (MD)}} 
& \multicolumn{2}{c|}{\textbf{Code-Contests}} 
& \multicolumn{2}{c}{\textbf{PWAB}} \\
\cmidrule{2-3} \cmidrule{4-5} \cmidrule{6-7}
& \textbf{LLaMA+\our} & \textbf{Qwen+\our}
& \textbf{LLaMA+\our} & \textbf{Qwen+\our}
& \textbf{LLaMA+\our} & \textbf{Qwen+\our} \\
\midrule
$(8.0, 1.0)$ & 4.00 & 4.15 & 2.11 & 2.81 & 3.96 & 3.71 \\[4pt]
$(2.0, 0.5)$ & 3.75 & 4.01 & 2.12 & 2.89 & 4.06 & 3.91 \\[4pt]
$(0.0, 0.2)$ & 3.70 & 3.91 & 2.08 & 2.79 & 4.17 & 3.80 \\
\bottomrule
\end{tabular}
}
\caption{
Tabular illustration of $\mu$Score variations for different $(\lambda_1,\lambda_2)$ values controlling missing dimension activation and diversity, respectively. Performance degrades as $\lambda$ decreases in Medical and PWAB, while Code-Contests remains comparatively stable, reflecting domain-specific sensitivity to proactive calibration.
}
\label{tab:lambda_calibration}
\end{table*}

While all three responses are factually plausible, they differ markedly in how implicit, task-relevant dimensions are addressed.  The evaluator model (\gpt) rated \our as the highest, noting: ``\textit{Provides diabetes-specific risk framing, vaccination context, practical precautions}''. The base \llama response focuses on general risk and transmission mechanisms and \gptf narrows the scope to likely outcomes and metabolic management but avoids broader risk framing or preventive action. In contrast, \our integrates multiple latent considerations—disease severity, glycemic control, vaccination, behavioral precautions, and comorbidities—into a single coherent response.

\begin{figure}[t!]
\centering
  \includegraphics[width=\columnwidth]{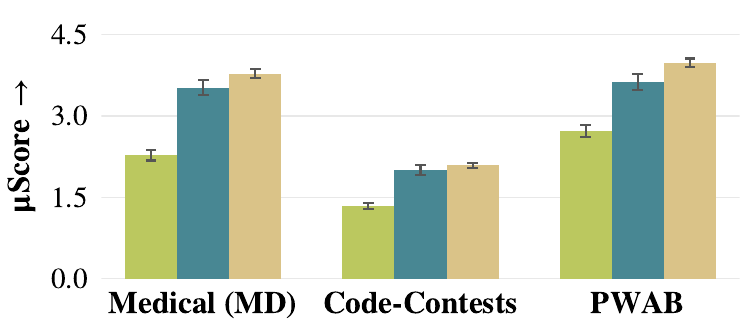}
  \includegraphics[height=0.6cm]{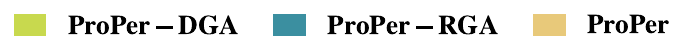}
\vspace{-0.3cm}
\caption{Performance comparison of \our and its variants (\our-DGA and \our-RGA), showing the impact of removing DGA or RGA on overall results. The drop without DGA is more pronounced.}
\label{fig:ablation}
\end{figure}

\begin{figure}[t!]
\centering
  \includegraphics[width=\columnwidth]{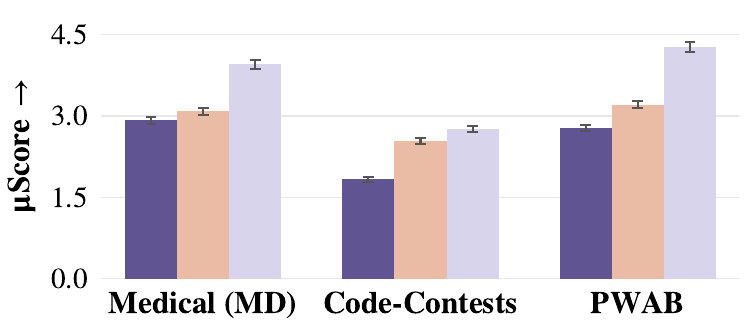}
  \includegraphics[height=0.7cm]{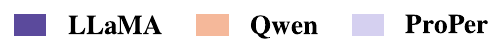}
\vspace{-0.3cm}
\caption{Quality comparison of dimensions generated by DGA in \our, \llama, and \qwen. \our produces the most effective dimensions, with \qwen outperforming \llama.}
\label{fig:dga}
\end{figure}

\xhdr{RQ2: Component Contributions.}
We analyze the contribution of \our's core components through targeted ablations. As shown in Figure~\ref{fig:ablation}, removing the Dimension Generating Agent (\textsc{DGA}) results in a substantial performance drop across all datasets, while removing post-hoc dimension reranking (\our-RGA) leads to a smaller but consistent degradation. This indicates that explicitly generating task-relevant implicit dimensions is foundational to \our's effectiveness: without \textsc{DGA}, the system lacks a structured representation of what is missing from the user query, and downstream response generation becomes less targeted and less calibrated.

To further understand this effect, Figure~\ref{fig:dga} compares the quality of dimensions produced by \textsc{DGA} against those generated directly by strong base LLMs. \textsc{DGA} consistently generates higher-quality dimensions than both \llama and \qwen, suggesting that the gains observed in the full system stem from learning to surface relevant latent needs rather than from response-generation capacity alone. Taken together, these results reveal a clear division of labor: \textsc{DGA} enables anticipation by identifying meaningful implicit dimensions, while reranking and \textsc{RGA} modulate their influence to prevent over-proactivity.

\xhdr{Proactivity vs.\ Personalization (RQ3)}
Table~\ref{tab:lambda_calibration} evaluates whether \our's gains arise from calibrated proactivity rather than indiscriminate response expansion by varying the calibration regime $(\lambda_1,\lambda_2)$ controlling implicit-dimension activation and diversity. In Medical and PWAB, reducing $(\lambda_1,\lambda_2)$ consistently lowers response quality for both backbones, indicating that under-activating implicit dimensions, such as safety considerations, latent constraints, and comparison criteria, leads to less helpful responses even when outputs remain concise and on-topic. In contrast, Code-Contests exhibits minimal sensitivity to calibration, reflecting a domain where objectives are tightly specified and additional proactive breadth offers limited benefit beyond producing a correct solution. Across all settings, the shared directionality of these trends for both \llama+\our and \qwen+\our suggests that the observed effects are not driven by model scale, but by the extent to which proactive dimension activation aligns with domain-specific task structure.
\label{sec:multi_turn_eval}

\begin{figure}[t]
    \centering
    \includegraphics[width=\columnwidth]{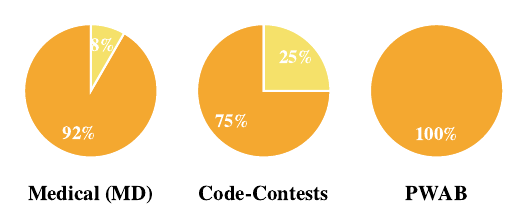}

    \centering
    \includegraphics[width=0.8\columnwidth]{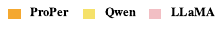}
    \vspace{-0.3cm}
    \caption{\our vs. Baseline: preference judgments on \gpt simulated multi-turn conversations (12 samples per domain).}
    \label{fig:pie_chart}
    \vspace{-0.3cm}
\end{figure}

\xhdr{On Multi-turn Conversations (RQ4)}
To assess whether calibrated proactivity remains stable over time, we conduct a small-scale multi-turn evaluation using 12 randomly sampled single-turn conversations per domain. Each sampled conversation is extended into a simulated multi-turn dialogue using GPT, and we then evaluate full interaction trajectories generated by \our against strong base-model baselines. As summarized in Figure~\ref{fig:pie_chart}, \our is preferred in the majority of conversations across domains (11/12 in Medical, 9/12 in Code-Contests, and 12/12 in PWAB). Baseline wins occur primarily in narrowly specified tasks where conservative responses suffice. These results suggest that \our's advantages extend beyond single-turn settings, maintaining appropriate initiative across turns, particularly in domains where needs and constraints emerge gradually.

\section{Conclusions and Discussion}
This work reframes proactive assistance as an epistemic calibration problem focused on identifying \emph{what is missing}. Rather than expanding responses, we treat proactivity as surfacing latent, task-relevant knowledge gaps — including needs users may not articulate. We introduce \our, a modular framework that models explicit and implicit dimensions to generate and prioritize proactive behavior in a principled way. Across three domains, \our improves task utility over strong baselines, with the largest gains where unarticulated risks, constraints, or trade-offs matter most. Ablations show that modeling unknowns is more critical than response-generation capacity alone, with benefits persisting in multi-turn settings.

More broadly, treating dimensions as intermediate representations raises questions about when to surface gaps, how to modulate initiative, and how to acknowledge uncertainty. This suggests directions for adaptive calibration, concept-grounded dimensions~\cite{sun2025conceptbottleneck}, and detecting incomplete representations — pointing toward agents that support not just goal execution, but sense-making by identifying unasked questions.

\section*{Acknowledgments}
The work described in this paper is supported by the US National Science Foundation (NSF) Award IIS-2336769. We utilized computational resources provided by the Hyak supercomputing system at the University of Washington (UW),  and was funded by the UW student technology fee.

\section{Limitations}

\paragraph{Evaluation choice and what it enables.}
Our goal in this paper is to isolate whether \emph{calibrated} activation of implicit dimensions improves assistance quality across domains. To make that comparison feasible at scale and under controlled conditions, we primarily use a strong LLM judge that applies a consistent rubric across many paired responses. This choice trades off some ecological validity for comparability: an LLM judge may reward certain writing styles or forms of reasoning, and it is not a substitute for human satisfaction or downstream task outcomes. We view this as a first step for studying calibration effects cleanly, and we expect the most valuable next step to be human-centered validation that measures trust, perceived intrusiveness, and long-horizon utility in addition to response quality.

\paragraph{Why dimensions are free-form (for now).}
We represent implicit dimensions in free-form language rather than grounding them in ontologies, taxonomies, or structured variables. This is deliberate: domain-specific grounding would hard-code priors that blur the causal question we study (does explicit implicit-dimension modeling help, independent of domain knowledge engineering?). The downside is reduced interpretability and weaker guarantees against redundancy or inconsistent phrasing. At the same time, the dimension interface creates a natural bridge to concept modeling: future variants could replace or augment textual dimensions with concept bottlenecks, ontology-aligned slots, or hybrid symbolic–neural representations, making proactivity more controllable and audit-able without sacrificing generality.

\paragraph{Calibration as a controlled knob rather than a learned policy.}
We sweep fixed $(\lambda_1,\lambda_2)$ regimes to expose how proactivity and diversity affect utility, instead of learning a policy that adapts these parameters online. We make this choice because fixed regimes yield clearer ablations and more interpretable conclusions about when proactivity helps versus when it saturates. The limitation is that real users likely require context- and user-dependent calibration. A natural next direction is to learn adaptive calibration that conditions on uncertainty, conversation stage, and user feedback, effectively turning $(\lambda_1,\lambda_2)$ into a dynamic control policy rather than static hyper-parameters.

\paragraph{Multi-turn evidence as a robustness check.}
Our multi-turn experiment is intentionally small-scale and qualitative. We use it to test a specific failure mode of proactive systems, whether initiative drifts or compounds across turns, rather than to claim comprehensive coverage of long-horizon behavior. Scaling this evaluation to longer trajectories with diverse interaction patterns is important, but it requires different study design and substantially more annotation. We see this as a key direction for advancing proactive agents: measuring not only per-turn helpfulness, but also stability of calibration, recovery from missteps, and user-perceived intrusiveness over time.

\paragraph{Personalization and modality beyond text.}
Although \our reasons over explicit and implicit dimensions, it does not yet maintain persistent user models (e.g., expertise, risk tolerance, or preference for initiative), and it operates purely over text. We adopt this restriction to stay compatible with widely used benchmarks and to keep the study focused on the proactivity mechanism itself. Extending the framework to user-specific proactivity thresholds and to multi-modal signals (behavioral traces, structured task state, or environment feedback) is a promising path to make calibration both more personalized and more robust in real deployment settings.

\paragraph{Toward unknown unknowns and epistemic proactivity.}
Finally, while \our surfaces missing task-relevant dimensions, it does not explicitly distinguish between known unknowns (recoverable gaps) and deeper epistemic gaps where the system may not even know what it is missing. We view this paper as a step toward that broader goal: by elevating dimensions into an explicit intermediate representation, we create a scaffold for future work to model epistemic uncertainty, detect when the dimension set itself is incomplete, and reason about when to ask, when to caution, and when to defer.

\paragraph{Limitations in high-stakes domains.}
While \our performs well in the Medical domain, deploying proactive assistance in high-stakes clinical settings raises considerations beyond benchmark accuracy. Miscalibrated interventions (particularly incorrectly inferred implicit dimensions) may carry unintended consequences for vulnerable users. Our evaluation does not assess safety or appropriateness in real clinical workflows, nor regulatory and ethical requirements of medical decision support. We leave safe proactive intervention in high-stakes domains to future work.
\bibliography{references}

@inproceedings{zargham2022understanding,
    author = {Zargham, Nima and Reicherts, Leon and Bonfert, Michael and Voelkel, Sarah Theres and Schoening, Johannes and Malaka, Rainer and Rogers, Yvonne},
    title = {Understanding Circumstances for Desirable Proactive Behaviour of Voice Assistants: The Proactivity Dilemma},
    year = {2022},
    booktitle = {Proceedings of the 4th Conference on Conversational User Interfaces},
    articleno = {3},
}

@inproceedings{
    li2025agentoriented,
    title={Agent-Oriented Planning in Multi-Agent Systems},
    author={Ao Li and Yuexiang Xie and Songze Li and Fugee Tsung and Bolin Ding and Yaliang Li},
    booktitle={The Thirteenth International Conference on Learning Representations},
    year={2025},
    url={https://openreview.net/forum?id=EqcLAU6gyU}
}

@article{cheng2025atsf,
  title={Position: Beyond Model-Centric Prediction--Agentic Time Series Forecasting},
  author={Cheng, Mingyue and Tao, Xiaoyu and Liu, Qi and Guo, Ze and Chen, Enhong},
  journal={arXiv preprint arXiv:2602.01776},
  year={2026}
}

@inproceedings{guptaproactive,
author = {Gupta, Vinayak and Bedathur, Srikanta},
title = {ProActive: Self-Attentive Temporal Point Process Flows for Activity Sequences},
year = {2022},
booktitle = {Proceedings of the 28th ACM SIGKDD Conference on Knowledge Discovery and Data Mining},
pages = {496–504}
}

@article{chen2025needhelp,
  title = {STYLE: Improving Domain Transferability of Asking Clarification Questions in Large Language Model Powered Conversational Agents},
  author = {Chen, Yue and Huang, Chen and Deng, Yang and Chua, Tat-Seng},
  journal = {Conference on Empirical Methods in Natural Language Processing (EMNLP)},
  year = {2025}
}

@inproceedings{diebel2025ai,
  title = {Trust Me, I’m Funny: Humor, Personalization, and Trust in Conversational Agents – A Systematic Literature Review on User Engagement, Educational Adoption, and Responsible Use},
  author = {Brîncoveanu, Constantin},
  booktitle = {Conference on Empirical Methods in Natural Language Processing (EMNLP)},
  year = {2025}
}

@article{pu2025assistance,
  title = {A Survey on Proactive Dialogue Systems: Problems, Methods, and Prospects},
  author = {Lei, Wenqiang and Chua, Tat-Seng and Deng, Yang and Lam, Wai},
  journal = {Conference on Empirical Methods in Natural Language Processing (EMNLP)},
  year = {2025},
}

@article{qwen3,
    title={Qwen3 Technical Report}, 
    author={An Yang and Anfeng Li and Baosong Yang and Beichen Zhang and Binyuan Hui and Bo Zheng and Bowen Yu and Chang Gao and Chengen Huang and Chenxu Lv and Chujie Zheng and Dayiheng Liu and Fan Zhou and Fei Huang and Feng Hu and Hao Ge and Haoran Wei and Huan Lin and Jialong Tang and Jian Yang and Jianhong Tu and Jianwei Zhang and Jianxin Yang and Jiaxi Yang and Jing Zhou and Jingren Zhou and Junyang Lin and Kai Dang and Keqin Bao and Kexin Yang and Le Yu and Lianghao Deng and Mei Li and Mingfeng Xue and Mingze Li and Pei Zhang and Peng Wang and Qin Zhu and Rui Men and Ruize Gao and Shixuan Liu and Shuang Luo and Tianhao Li and Tianyi Tang and Wenbiao Yin and Xingzhang Ren and Xinyu Wang and Xinyu Zhang and Xuancheng Ren and Yang Fan and Yang Su and Yichang Zhang and Yinger Zhang and Yu Wan and Yuqiong Liu and Zekun Wang and Zeyu Cui and Zhenru Zhang and Zhipeng Zhou and Zihan Qiu},
    journal = {arXiv preprint arXiv:2505.09388},
    year={2025}
}

@inproceedings{cai2024personalwab,
      title={Large Language Models Empowered Personalized Web Agents}, 
      author={Hongru Cai and Yongqi Li and Wenjie Wang and Fengbin Zhu and Xiaoyu Shen and Wenjie Li and Tat-Seng Chua},
      year={2025},
      booktitle={Proceedings of the ACM Web Conference 2025},
      series={WWW '25}
}

@inproceedings{jiang2024unknown,
  title     = {Into the Unknown Unknowns: Engaged Human Learning through Participation in Language Model Agent Conversations},
  author    = {Jiang, Yucheng and Shao, Yijia and Ma, Dekun and Semnani, Sina and Lam, Monica},
  booktitle = {Proceedings of the 2024 Conference on Empirical Methods in Natural Language Processing},
  pages     = {9917--9955},
  year      = {2024},
  address   = {Miami, Florida, USA},
  publisher = {Association for Computational Linguistics}
}

@inproceedings{proactive_recs,
	author = {Bi, Shuxian and Wang, Wenjie and Pan, Hang and Feng, Fuli and He, Xiangnan},
	title = {Proactive Recommendation with Iterative Preference Guidance},
	year = {2024},
	url = {https://doi.org/10.1145/3589335.3651548},
	doi = {10.1145/3589335.3651548},
	booktitle = {Companion Proceedings of the ACM Web Conference 2024},
	pages = {871–874},
	numpages = {4},
}

@article{gupta2022doing,
  title={Doing more with less: overcoming data scarcity for poi recommendation via cross-region transfer},
  author={Gupta, Vinayak and Bedathur, Srikanta},
  journal={ACM Transactions on Intelligent Systems and Technology (TIST)},
  volume={13},
  number={3},
  pages={1--24},
  year={2022},
  publisher={ACM New York, NY}
}

@article{kaur2026efficient,
  title={Efficient and responsible adaptation of large language models for robust top-k recommendations},
  author={Kaur, Kirandeep and Gupta, Vinayak and Chadha, Manya and Shah, Chirag},
  journal={ACM Transactions on Recommender Systems},
  volume={4},
  number={3},
  pages={1--31},
  year={2026},
  publisher={ACM New York, NY}
}

@article{kaur2026knowing,
  title={Knowing Isn't Understanding: Re-grounding Generative Proactivity with Epistemic and Behavioral Insight},
  author={Kaur, Kirandeep and Lyu, Xingda and Shah, Chirag},
  journal={arXiv preprint arXiv:2602.15259},
  year={2026}
}

@article{gupta_proper_tist,
author = {Gupta, Vinayak and Bedathur, Srikanta},
title = {Tapestry of Time and Actions: Modeling Human Activity Sequences Using Temporal Point Process Flows},
year = {2024},
issue_date = {June 2024},
publisher = {Association for Computing Machinery},
address = {New York, NY, USA},
volume = {15},
number = {3},
doi = {10.1145/3650045},
journal = {ACM Trans. Intell. Syst. Technol.},
articleno = {49},
}

@inproceedings{li2025knowledge,
  title     = {Knowledge Boundary of Large Language Models: A Survey},
  author    = {Li, Moxin and Zhao, Yong and Zhang, Wenxuan and Li, Shuaiyi and Xie, Wenya and Ng, See-Kiong and Chua, Tat-Seng and Deng, Yang},
  booktitle = {Proceedings of the 63rd Annual Meeting of the Association for Computational Linguistics (Volume 1: Long Papers)},
  pages     = {5131--5157},
  year      = {2025},
  address   = {Vienna, Austria},
  publisher = {Association for Computational Linguistics}
}

@article{griot2025metacognition,
  title   = {Large Language Models lack essential metacognition for reliable medical reasoning},
  author  = {Griot, M. and Hemptinne, C. and Vanderdonckt, J. and others},
  journal = {Nature Communications},
  volume  = {16},
  pages   = {642},
  year    = {2025},
  doi     = {10.1038/s41467-024-55628-6}
}

@article{li2022competition,
  title={Competition-Level Code Generation with AlphaCode},
    author={Li, Yujia and Choi, David and Chung, Junyoung and Kushman, Nate and
    Schrittwieser, Julian and Leblond, R{\'e}mi and Eccles, Tom and
    Keeling, James and Gimeno, Felix and Dal Lago, Agustin and
    Hubert, Thomas and Choy, Peter and de Masson d'Autume, Cyprien and
    Babuschkin, Igor and Chen, Xinyun and Huang, Po-Sen and Welbl, Johannes and
    Gowal, Sven and Cherepanov, Alexey and Molloy, James and
    Mankowitz, Daniel and Sutherland Robson, Esme and Kohli, Pushmeet and
    de Freitas, Nando and Kavukcuoglu, Koray and Vinyals, Oriol},
  journal={arXiv preprint arXiv:2203.07814},
  year={2022}
}

@article{Bandyopadhyay2025YETIT,
  title={YETI (YET to Intervene) Proactive Interventions by Multimodal AI Agents in Augmented Reality Tasks},
  author={Saptarashmi Bandyopadhyay and Vikas Bahirwani and Lavisha Aggarwal and Bhanu Prakash Reddy Guda and Lin Li and Andrea Colaco},
  journal={ArXiv},
  year={2025},
  volume={abs/2501.09355},
  url={https://api.semanticscholar.org/CorpusID:275570341}
}

@inproceedings{dong-etal-2025-protod,
    title = "{P}ro{TOD}: Proactive Task-oriented Dialogue System Based on Large Language Model",
    author = "Dong, Wenjie  and
      Chen, Sirong  and
      Yang, Yan",
    editor = "Rambow, Owen  and
      Wanner, Leo  and
      Apidianaki, Marianna  and
      Al-Khalifa, Hend  and
      Eugenio, Barbara Di  and
      Schockaert, Steven",
    booktitle = "Proceedings of the 31st International Conference on Computational Linguistics",
    month = jan,
    year = "2025",
    address = "Abu Dhabi, UAE",
    publisher = "Association for Computational Linguistics",
    url = "https://aclanthology.org/2025.coling-main.614/",
    pages = "9147--9164",
}

@inproceedings{
    hahn2025proactive,
    title={Proactive Agents for Multi-Turn Text-to-Image Generation Under Uncertainty},
    author={Meera Hahn and Wenjun Zeng and Nithish Kannen and Rich Galt and Kartikeya Badola and Been Kim and Zi Wang},
    booktitle={Forty-second International Conference on Machine Learning},
    year={2025},
    url={https://openreview.net/forum?id=f3iBgm2Zi0}
}

@inproceedings{10.1609/aaai.v38i16.29710,
author = {Zhang, Ceyao and Yang, Kaijie and Hu, Siyi and Wang, Zihao and Li, Guanghe and Sun, Yihang and Zhang, Cheng and Zhang, Zhaowei and Liu, Anji and Zhu, Song-Chun and Chang, Xiaojun and Zhang, Junge and Yin, Feng and Liang, Yitao and Yang, Yaodong},
title = {ProAgent: building proactive cooperative agents with large language models},
year = {2024},
isbn = {978-1-57735-887-9},
publisher = {AAAI Press},
url = {https://doi.org/10.1609/aaai.v38i16.29710},
doi = {10.1609/aaai.v38i16.29710},
booktitle = {Proceedings of the Thirty-Eighth AAAI Conference on Artificial Intelligence and Thirty-Sixth Conference on Innovative Applications of Artificial Intelligence and Fourteenth Symposium on Educational Advances in Artificial Intelligence},
articleno = {1962},
numpages = {9},
series = {AAAI'24/IAAI'24/EAAI'24}
}

@inproceedings{10.1145/3626772.3657843,
author = {Deng, Yang and Liao, Lizi and Zheng, Zhonghua and Yang, Grace Hui and Chua, Tat-Seng},
title = {Towards Human-centered Proactive Conversational Agents},
year = {2024},
isbn = {9798400704314},
publisher = {Association for Computing Machinery},
address = {New York, NY, USA},
url = {https://doi.org/10.1145/3626772.3657843},
doi = {10.1145/3626772.3657843},
booktitle = {Proceedings of the 47th International ACM SIGIR Conference on Research and Development in Information Retrieval},
pages = {807–818},
numpages = {12},
location = {Washington DC, USA},
series = {SIGIR '24}
}

@inproceedings{deng-etal-2024-dont,
    title = "Don`t Just Say {\textquotedblleft}{I} don`t know{\textquotedblright}! Self-aligning Large Language Models for Responding to Unknown Questions with Explanations",
    author = "Deng, Yang  and
      Zhao, Yong  and
      Li, Moxin  and
      Ng, See-Kiong  and
      Chua, Tat-Seng",
    editor = "Al-Onaizan, Yaser  and
      Bansal, Mohit  and
      Chen, Yun-Nung",
    booktitle = "Proceedings of the 2024 Conference on Empirical Methods in Natural Language Processing",
    month = nov,
    year = "2024",
    address = "Miami, Florida, USA",
    publisher = "Association for Computational Linguistics",
    url = "https://aclanthology.org/2024.emnlp-main.757/",
    doi = "10.18653/v1/2024.emnlp-main.757",
    pages = "13652--13673",
}

@inproceedings{zhang-etal-2024-ask,
    title = "Ask-before-Plan: Proactive Language Agents for Real-World Planning",
    author = "Zhang, Xuan  and
      Deng, Yang  and
      Ren, Zifeng  and
      Ng, See-Kiong  and
      Chua, Tat-Seng",
    editor = "Al-Onaizan, Yaser  and
      Bansal, Mohit  and
      Chen, Yun-Nung",
    booktitle = "Findings of the Association for Computational Linguistics: EMNLP 2024",
    month = nov,
    year = "2024",
    address = "Miami, Florida, USA",
    publisher = "Association for Computational Linguistics",
    url = "https://aclanthology.org/2024.findings-emnlp.636/",
    doi = "10.18653/v1/2024.findings-emnlp.636",
    pages = "10836--10863"
}

@article{none_too_solid,
author = {Ann Kerwin},
title ={None Too Solid: Medical Ignorance},
journal = {Knowledge},
volume = {15},
number = {2},
pages = {166-185},
year = {1993},
doi = {10.1177/107554709301500204},
URL = {https://doi.org/10.1177/107554709301500204},
eprint = {https://doi.org/10.1177/107554709301500204},
}

@inproceedings{deng-etal-2023-prompting,
    title = "Prompting and Evaluating Large Language Models for Proactive Dialogues: Clarification, Target-guided, and Non-collaboration",
    author = "Deng, Yang  and
      Liao, Lizi  and
      Chen, Liang  and
      Wang, Hongru  and
      Lei, Wenqiang  and
      Chua, Tat-Seng",
    editor = "Bouamor, Houda  and
      Pino, Juan  and
      Bali, Kalika",
    booktitle = "Findings of the Association for Computational Linguistics: EMNLP 2023",
    month = dec,
    year = "2023",
    address = "Singapore",
    publisher = "Association for Computational Linguistics",
    url = "https://aclanthology.org/2023.findings-emnlp.711/",
    doi = "10.18653/v1/2023.findings-emnlp.711",
    pages = "10602--10621",
}

@inproceedings{
lu2024proactiveagent,
title={Proactive Agent: Shifting {LLM} Agents from Reactive Responses to Active Assistance},
author={Yaxi Lu and Shenzhi Yang and Cheng Qian and Guirong Chen and Qinyu Luo and Yesai Wu and Huadong Wang and Xin Cong and Zhong Zhang and Yankai Lin and Weiwen Liu and Yasheng Wang and Zhiyuan Liu and Fangming Liu and Maosong Sun},
booktitle={The Thirteenth International Conference on Learning Representations},
year={2025},
url={https://openreview.net/forum?id=sRIU6k2TcU}
}

@article{madaan2023self,
  title={Self-refine: Iterative refinement with self-feedback},
  author={Madaan, Aman and Tandon, Niket and Gupta, Prakhar and Hallinan, Skyler and Gao, Luyu and Wiegreffe, Sarah and Alon, Uri and Dziri, Nouha and Prabhumoye, Shrimai and Yang, Yiming and others},
  journal={Advances in Neural Information Processing Systems},
  volume={36},
  pages={46534--46594},
  year={2023}
}

@article{grattafiori2024llama,
  title={The llama 3 herd of models},
  author={Grattafiori, Aaron and Dubey, Abhimanyu and Jauhri, Abhinav and Pandey, Abhinav and Kadian, Abhishek and Al-Dahle, Ahmad and Letman, Aiesha and Mathur, Akhil and Schelten, Alan and Vaughan, Alex and others},
  journal={arXiv preprint arXiv:2407.21783},
  year={2024}
}

@article{Liu2024ProactiveCA,
  title={Proactive Conversational Agents with Inner Thoughts},
  author={Xingyu Bruce Liu and Shitao Fang and Weiyan Shi and Chien-Sheng Wu and Takeo Igarashi and Xiang 'Anthony' Chen},
  journal={Proceedings of the 2025 CHI Conference on Human Factors in Computing Systems},
  year={2024},
  url={https://api.semanticscholar.org/CorpusID:275213148}
}

@inproceedings{Huang2024PromptingEA,
  title={Prompting Explicit and Implicit Knowledge for Multi-hop Question Answering Based on Human Reading Process},
  author={Guangming Huang and Yunfei Long and Cunjin Luo and Jiaxing Shen and Xia Sun},
  booktitle={International Conference on Language Resources and Evaluation},
  year={2024},
  url={https://api.semanticscholar.org/CorpusID:268091260}
}

@misc{llamafactory2024,
  title = {LLaMA-Factory: Unified Efficient Fine-Tuning of Large Language Models},
  author = {Zheng, Yaowei and others},
  year = {2024},
  howpublished = {\url{https://github.com/hiyouga/LLaMA-Factory}}
}

@inproceedings{medDG,
author = {Liu, Wenge and Tang, Jianheng and Cheng, Yi and Li, Wenjie and Zheng, Yefeng and Liang, Xiaodan},
title = {MedDG: An Entity-Centric Medical Consultation Dataset for Entity-Aware Medical Dialogue Generation},
year = {2022},
isbn = {978-3-031-17119-2},
publisher = {Springer-Verlag},
address = {Berlin, Heidelberg},
url = {https://doi.org/10.1007/978-3-031-17120-8_35},
doi = {10.1007/978-3-031-17120-8_35},
booktitle = {Natural Language Processing and Chinese Computing: 11th CCF International Conference, NLPCC 2022, Guilin, China, September 24–25, 2022, Proceedings, Part I},
pages = {447–459},
numpages = {13},
location = {Guilin, China}
}

@inproceedings{park2024generativeagents,
  title     = {Generative Agents: Interactive Simulacra of Human Behavior},
  author    = {Park, Joon Sung and O'Brien, Joseph C. and Cai, Carrie J. and Morris, Meredith Ringel and Liang, Percy and Bernstein, Michael S.},
  booktitle = {Proceedings of the International Conference on Learning Representations (ICLR)},
  year      = {2024},
  url       = {https://arxiv.org/abs/2304.03442}
}

@inproceedings{ribeiro2020beyondaccuracy,
  title     = {Beyond Accuracy: Behavioral Testing of NLP Models with CheckList},
  author    = {Ribeiro, Marco Tulio and Wu, Tongshuang and Guestrin, Carlos and Singh, Sameer},
  booktitle = {Proceedings of the 58th Annual Meeting of the Association for Computational Linguistics (ACL)},
  year      = {2020},
  pages     = {4902--4912},
  url       = {https://aclanthology.org/2020.acl-main.442/}
}

@inproceedings{sun2025conceptbottleneck,
  title     = {Concept Bottleneck Large Language Models},
  author    = {Sun, Chung-En and Lee, Sungjin and Zhang, Yiyang and Chen, Danqi},
  booktitle = {Proceedings of the International Conference on Learning Representations (ICLR)},
  year      = {2025},
  url       = {https://arxiv.org/abs/2407.10856}
}

\clearpage
\appendix
\begin{center}
\large \textbf{Supplementary Material for: \papertitle}
\end{center}

\section{Fine-tuning Data Construction Pipeline}
\label{app:data_pipeline}
To enable models to surface missing knowledge gaps, we construct fine-tuning data at the level of \emph{interaction dimensions} that goes beyond raw text and elicits major aspects and considerations unique to an interaction that guide the successful completion of a certain task. Dimension-level supervision abstracts away surface form variation and allows the DGA to learn patterns of well-articulated user needs and task-relevant aspects considered to solve that task. Below we describe how such dimension-level supervision is defined, generated, and used to fine-tune the DGA.

\subsection{Dimension Annotation Schema}
\label{app:dimension_schema}

This subsection defines the annotation schema used to construct supervision for training the Dimension Generating Agent (DGA). Importantly, our DGA is trained only on \emph{observed} dimensions i.e., dimensions that are explicitly present in the user input and/or explicitly addressed in the assistant response, to learn what aspects are salient for a given type of task. At inference time, we then prompt the DGA to \emph{elicit implicit dimensions} by proposing which salient dimensions are likely missing from the current interaction, conditioned on the user state.

We represent each interaction using a shared universe of dimensions, where a dimension is a short descriptor of a task-relevant aspect that may influence response quality (e.g., ``input validation,'' ``safety constraints,'' ``preference trade-offs''). For a given interaction state, we annotate two explicit sets. \emph{User-explicit dimensions} capture aspects explicitly stated or clearly implied by the user query or interaction history. \emph{System-explicit dimensions} capture aspects addressed by a baseline (or reference) assistant response. These explicit annotations provide the DGA with dimension-level signals of what matters for similar tasks and how such aspects tend to be expressed or addressed.

Implicit dimensions are not directly supervised during fine-tuning. Instead, they are elicited at inference by conditioning the DGA on the current interaction and prompting it to propose task-relevant dimensions that are likely missing from the user input and baseline response, leveraging the dimension vocabulary and regularities learned from explicit supervision.

\subsection{Fine-tuning Data Generation Procedure}
\label{app:finetuning_pipeline}
While the concrete data generation procedures differ across datasets and domains, all fine-tuning data for the Dimension Generating Agent (DGA) follows a shared high-level pipeline. This pipeline defines how raw interactions are transformed into structured, domain-independent supervision at the dimension level.

At a high level, the process begins with a user query and the corresponding assistant response observed in an existing interaction. Given this interaction context, we use \textbf{GPT-5} and apply a curated elicitation prompt designed to surface task-relevant dimensions expressed on the user side and dimensions addressed by the assistant response. These prompts are tailored to the characteristics of each dataset and domain, and are described in detail in subsequent subsections.The elicitation prompt produces a structured output in the form of a JSON object, containing the inferred user-explicit dimensions and system-explicit dimensions for the interaction. These dimension annotations are then stored and aggregated to form the fine-tuning dataset. Each training example thus consists of an interaction context paired with a structured representation of explicit dimensions, abstracting away surface-level textual variation.
In the following subsections, we describe the concrete instantiations of this pipeline for each dataset, including the specific prompts used and any domain-specific heuristics or filtering applied during data construction.

\noindent(i)~\textbf{CodeContest Annotation Pipeline:}
We first instantiate the dimension annotation pipeline on the CodeContest dataset, which consists of competitive programming problems and corresponding reference solutions. Since CodeContest does not natively contain user queries, we construct realistic user–assistant interaction contexts by eliciting user-style queries from problem descriptions. This enables downstream dimension annotation while preserving the characteristics of real coding assistance scenarios.

\noindent\textbf{Initial Data Preprocessing.}
We begin with raw CodeContest data dumps containing programming problems and associated metadata. All available splits are concatenated into a unified dataset comprising 13{,}610 problem instances. Each instance includes attributes such as problem description, difficulty, test cases, and multiple reference solutions.

For the scope of our work, we retain only the following fields:
\begin{itemize}
    \item \textbf{Problem Description}: Natural language description of the programming task.
    \item \textbf{Difficulty}: Integer-coded difficulty level.
    \item \textbf{Solutions}: A list of reference code solutions with language annotations.
\end{itemize}

We further restrict the dataset to \textbf{Python~3} solutions to ensure consistency and ease of downstream processing, as Python~3 is the most prevalent and actively maintained language in the dataset.

\noindent\textbf{Difficulty-Aware Dataset Construction.}
After filtering, the dataset spans 18 distinct difficulty levels. To evaluate both in-distribution learning and generalization, we partition the dataset into \emph{warm} and \emph{cold-start} subsets. The warm set includes up to 15 problems per difficulty level (or all problems when fewer are available), and is used for training and validation. The remaining problems constitute the cold-start set, enabling evaluation on entirely unseen problem instances.

For each problem, we retain at most 50 Python~3 solutions. Each entry is flattened into the following structure:

{
\small
\[
\langle \text{Problem Description},\ \text{Difficulty},\ \text{Python~3 Solutions} \rangle
\]
}

The warm subset is randomly split into training and test sets using a 70/30 ratio.

\noindent\textbf{Query Elicitation via Prompting.}
Because CodeContest does not include user-issued queries, we use a large language model (GPT-5) to elicit realistic user prompts conditioned on the problem description. The goal is not to generate solutions, but to simulate how users with different levels of programming experience might articulate their intent when interacting with a coding assistant.

Specifically, for each problem, we generate three query variants corresponding to increasing levels of user expertise. This design allows the annotation pipeline to capture variation in how explicit or underspecified user intent may be expressed.

\begin{table}[t]
\centering
\small
\resizebox{\columnwidth}{!}{
\begin{tabular}{p{2.2cm} p{5.5cm}}
\toprule
\textbf{Query Level} & \textbf{Description} \\
\midrule
Level~1 & Vague prompts with minimal technical detail, reflecting beginner-level understanding. \\
Level~2 & Moderately structured prompts with partial technical awareness and some ambiguity. \\
Level~3 & Precise, code-focused prompts reflecting strong understanding of algorithms or data structures. \\
\bottomrule
\end{tabular}
}
\caption{Query elicitation levels used to simulate user expertise in CodeContest.}
\label{tab:codecontest_query_levels}
\end{table}

\noindent\textbf{Prompt Templates.}
We use carefully designed prompt templates to elicit these query variants. Each prompt conditions on the problem description and instructs the model to generate a single realistic user query matching the desired expertise level. The prompts are shown below for completeness.

\vspace{0.5em}
\noindent\textbf{Level~1 (Beginner) Query Prompt}
\begin{quote}
\small
\ttfamily
You have been given a detailed description of a coding problem and its solution.
Your task is to generate one realistic user prompt that a beginner programmer might naturally ask GitHub Copilot to generate similar code.

Requirements:\\
- Natural phrasing that sounds like a real beginner developer.\\
- Very vague understanding of the problem.\\
- General intent without clear technical details.\\
- Concise and Copilot-friendly.

Input:
<problem description>

Output:
A single vague, beginner-level user prompt a programmer would use.
\end{quote}

\vspace{0.5em}
\noindent\textbf{Level~2 (Intermediate) Query Prompt}
\begin{quote}
\small
\ttfamily
You have been given a detailed description of a coding problem and its solution.
Your task is to generate one realistic user prompt that an intermediate programmer might naturally ask GitHub Copilot to generate similar code.

Requirements:\\
- Natural phrasing reflecting partial understanding.\\
- References to basic techniques (e.g., loops, conditionals, functions).\\
- Some structure, but remaining ambiguity in solution design.\\
- Clear and concise.

Input:
<problem description>

Output:
A single moderately clear user prompt showing partial understanding of the task.
\end{quote}

\vspace{0.5em}
\noindent\textbf{Level~3 (Advanced) Query Prompt}
\begin{quote}
\small
\ttfamily
You have been given a detailed description of a coding problem and its solution.
Your task is to generate one realistic user prompt that an experienced programmer might naturally ask GitHub Copilot to generate similar code.

Requirements:\\
- Precise, code-focused phrasing.\\
- Explicit references to relevant data structures, algorithms, or optimizations.\\
- Highly targeted intent guiding the solution approach.\\
- Concise and technically specific.

Input:
<problem description>

Output:
A single clear and specific user prompt capturing all relevant solution details.
\end{quote}

\noindent\textbf{Structured Annotation Output.}
The generated queries, along with the original problem descriptions and reference solutions, are passed to dataset-specific elicitation prompts (described in subsequent subsections) that infer user-explicit and system-explicit dimensions. The resulting annotations are stored as structured JSON objects and form the basis of the fine-tuning dataset used to train the Dimension Generating Agent.

\noindent\textbf{Illustrative Query Variants.}
The elicited queries differ substantially in how explicitly users articulate goals, constraints, and uncertainty. Table~\ref{tab:codecontest_query_examples} shows representative examples from the three query levels used in CodeContest. These variations allow the annotation pipeline to capture differences in how task intent and missing information are expressed across user expertise levels.

\begin{table}[t]
\centering
\small
\resizebox{\columnwidth}{!}{
\begin{tabular}{p{2.2cm} p{5.25cm}}
\toprule
\textbf{Query Level} & \textbf{Example Excerpt} \\
\midrule
Beginner & ``How can I write a program to find the minimal number of groups where the words in each group denote the same name given a list of words where some letters can be replaced by others and still be considered the same name?'' \\
Intermediate & ``Given a string s of length at most 1000 composed of lowercase letters, find whether it contains heidi as a subsequence, return YES or NO. Constraints: s is nonempty. Goal is to determine if s contains heidi as a subsequence without looking for heidi directly. Tricky part: Handle the case where heidi is a substring of s. Should I implement a function to check if s contains heidi as a subsequence or directly compare s and heidi?'' \\
Advanced & ``How to efficiently group words into equivalence classes under certain transformations with respect to the Latin alphabet, allowing for u to oo and h to kh replacements, while minimizing the number of groups, considering a large input size up to 400 words and time complexity of O(n*m) or better, where n is the number of words and m is the maximum word length, ensuring correct handling of edge cases such as single-letter words and empty strings?'' \\
\bottomrule
\end{tabular}
}
\caption{Representative query excerpts illustrating variation in specificity and technical detail across elicited expertise levels.}
\label{tab:codecontest_query_examples}
\end{table}

\noindent\textbf{Dimension Extraction via Prompted Annotation.}
Given a generated query and a reference solution, we next extract structured interaction dimensions using a curated annotation prompt. This prompt is designed to infer \emph{explicit user-side dimensions} and \emph{system-side dimensions} grounded in the solution code, without speculative reasoning.

The prompt instructs the model to reason along interpretable axes such as user intent, uncertainty, task structure, and constraints on the user side, and algorithmic patterns, complexity, and robustness on the system side. Importantly, the prompt requires all extracted dimensions to be justified using evidence from the input text, producing structured JSON outputs suitable for downstream processing.

This prompt design serves two purposes. First, it exposes the Dimension Generating Agent to consistent, semantically meaningful signals about what aspects are explicitly present in real coding interactions, despite wide variation in surface phrasing. Second, by anchoring extraction to both the query and the solution, it allows the model to internalize which technical dimensions tend to matter for different classes of programming tasks. These learned regularities later enable the DGA to elicit plausible \emph{implicit} dimensions at inference time when such aspects are missing from a new interaction.

\noindent\textbf{Annotation Prompt.}
For completeness, we show the annotation prompt used to extract dimensions below.
\begin{quote}
\small
\ttfamily{
You are an expert annotator for programming tasks. Extract latent dimensions from: \\
(1) User query, (2) Solution code. \\

Objectives: \\
- Infer concise, latent aspects grounded in the text. \\
- Avoid speculation beyond weak, text-supported inference. \\

User-side axes: \\
goal intent, user state, expertise level, uncertainty indicators, \\
constraint awareness, task structure, missing information. \\

System-side axes: \\
algorithm pattern, data structures, time and space complexity, \\
optimization decisions, robustness, constraint handling. \\

Output format (JSON only): \\
\{ \\
\ \ "user\_aspects": [ \\
\ \ \ \ \{"name": str, "value": str, "justification": str\} \\
\ \ ], \\
\ \ "solution\_aspects": [ \\
\ \ \ \ \{"name": str, "value": str, "justification": str\} \\
\ \ ] \\
\} \\
Justifications must cite evidence from the input or be null if implicit.
}
\end{quote}

\noindent(ii)~\textbf{Medical Diagnosis (MD) Annotation Pipeline:}
We next instantiate our annotation pipeline on the Medical Diagnosis (MD) dataset, which consists of $280$ clinical question–answer interactions between patients and medical professionals. We also apply our annotation pipeline to the Medical Diagnosis (MD) dataset, which consists of real-world clinical question--answer interactions between patients and medical professionals. Unlike CodeContest, this dataset natively contains user queries and expert responses, making it directly suitable for dimension-level annotation without additional preprocessing or query elicitation.

The MD dataset captures medically grounded interactions that include symptom descriptions, clinical uncertainty, risk considerations, and professional guidance. As a result, we directly operate on the original patient utterances and doctor responses. Additional details about the dataset are available at the public release link.\footnote{\url{https://drive.google.com/drive/folders/11sglwm6-cY7gjeqlZaMxL_MDKDMLdhym}}

\paragraph{Illustrative Interaction Example.}
Table~\ref{tab:md_example} shows a representative interaction from the MD dataset. Such examples highlight how patient queries often combine explicit symptoms with uncertainty and concern, while doctor responses balance reassurance, risk assessment, and guidance.

\begin{table}[t]
\centering
\small
\resizebox{\columnwidth}{!}{
\begin{tabular}{p{2.2cm} p{5.5cm}}
\toprule
\textbf{Speaker} & \textbf{Utterance} \\
\midrule
Patient & \emph{``I have had cold symptoms for over a week and a low-grade fever last week. For the past two days I have been feeling dizzy. Should I contact my doctor?''} \\
Doctor & \emph{``These symptoms alone are not enough to classify you as a COVID-19 suspect. However, if you have had contact with a confirmed case, experience persistent cough or shortness of breath, or have chronic conditions such as diabetes or immune suppression, you should seek medical attention and follow protective measures.''} \\
\bottomrule
\end{tabular}
}
\caption{Example patient--doctor interaction from the MD dataset.}
\label{tab:md_example}
\end{table}

\paragraph{Dimension Extraction Prompt.}
For each interaction, we apply a curated annotation prompt that extracts explicit and latent dimensions from both the patient query and the doctor response. The prompt emphasizes medically meaningful distinctions, including symptom patterns, risk indicators, and uncertainty handling, while explicitly discouraging speculative inference. The resulting structured annotations are used as supervision for training the Dimension Generating Agent.

\begin{quote}
\small
\ttfamily{
You are an expert annotator analyzing clinical question--answer interactions. \\
Extract both explicit and latent dimensions from: \\
(1) User medical query, (2) Doctor response. \\

Objectives: \\
- Infer concise, latent, and non-trivially stated aspects. \\
- Avoid speculation beyond weak, text-grounded inference. \\
- Emphasize medically relevant, domain-specific distinctions. \\

User-side axes: \\
clinical goal intent, explicit symptoms and history, latent symptom patterns, \\
disease-specific indicators, multisystem interactions, risk or red-flag indicators, \\
constraints, missing information, emotional state, task structure. \\

System-side axes: \\
medical reasoning patterns, diagnostic hypotheses, treatment guidance, \\
risk assessment, uncertainty handling, reassurance strategies, guideline alignment. \\

Output format (JSON only): \\
\{ \\
\ \ "user\_aspects": [ \\
\ \ \ \ \{"name": str, "value": str, "justification": str\} \\
\ \ ], \\
\ \ "solution\_aspects": [ \\
\ \ \ \ \{"name": str, "value": str, "justification": str\} \\
\ \ ] \\
\} \\

Justifications must cite evidence from the input or be null if implicit.
}
\end{quote}

\noindent(iii)~\textbf{PersonalWAB Dataset}
\label{app:personalwab_pipeline}

We further instantiate our annotation pipeline on the PersonalWAB dataset, which captures personalized shopping recommendation interactions grounded in user personas, historical behavior, and product metadata. Unlike CodeContest and MD, PersonalWAB explicitly models long-term user characteristics, making it well-suited for studying how personalization and proactivity interact in recommendation settings.

The dataset consists of four primary components: user profiles, user shopping history, user-issued instructions (queries), and a large catalog of products spanning multiple categories. Each interaction links a user persona and query to a recommended product, providing a rich context for extracting both preference-driven and constraint-driven dimensions. We follow the dataset's standard preprocessing pipeline, which includes consolidating user instructions, aligning them with corresponding user profiles, and filtering profiles to ensure balanced coverage across demographic attributes and occupations. All remaining profiles outside the curated subset are retained as a cold-start set for out-of-distribution evaluation.

\paragraph{Illustrative Interaction Structure.}
Each annotated instance in PersonalWAB (Table~\ref{tab:personalwab_example}) is defined by three inputs: (i) a user persona capturing long-term preferences and traits, (ii) a user query expressing a shopping intent, and (iii) a recommended product with structured attributes such as category, price, and features. This structure allows the annotation process to distinguish between stable user preferences, situational needs, and product-side affordances.

\begin{table}[t]
\centering
\small
\resizebox{\columnwidth}{!}{
\begin{tabular}{p{2.3cm} p{5.4cm}}
\toprule
\textbf{Component} & \textbf{Content} \\
\midrule
User Persona &
Gender: Male; Age: 35--44; Occupation: Technician/Engineer; \newline
Price Sensitivity: Medium; Interaction Complexity: Medium; \newline
Shopping Interests: Electronics, Home Improvement, Health \& Fitness; \newline
Brand Preferences: Apple, Sony, Under Armour, Duracell; \newline
Tone: Practical, straightforward, enthusiastic; \newline
Focus Aspects: Average rating, number of ratings, quality, durability
\\
\midrule
User Query &
``Hey there! I'm looking for some durable athletic accessories or gear.
Any recommendations for quality products with solid ratings?
Eager to explore some options!''
\\
\midrule
Recommended Product &
Title: Cramer Tuf-Skin Taping Base for Athletic Tape; \newline
Category: Athletic Tapes \& Wraps; \newline
Store: Cramer; \newline
Average Rating: 4.6; \newline
Review Highlight: Improves tape durability, water-resistant, easy removal
\\
\bottomrule
\end{tabular}
}
\caption{Example annotated interaction from the PersonalWAB dataset, illustrating the persona--query--product structure used for dimension extraction.}
\label{tab:personalwab_example}
\end{table}

\paragraph{Dimension Extraction Prompt.}
To extract interaction dimensions, we apply a curated annotation prompt that jointly reasons over the user persona, query, and recommended product. The prompt is designed to surface explicit preference signals as well as latent needs and constraints, while avoiding speculative inference beyond what is weakly supported by the input text. On the system side, extraction is restricted to properties of the recommended product itself.

\begin{quote}
\small
\ttfamily{
You are an expert annotator analyzing user--system interactions in shopping recommendation tasks. \\
Extract all the explicitly stated and latent dimensions from: \\
$(1)$ User persona, $(2)$ User query, $(3)$ Recommended product. \\

Objectives: \\
- Infer concise, latent, and non-trivially stated aspects. \\
- Avoid speculation beyond weak, text-grounded inference. \\

User-side axes: \\
goal intent, preference profile, need-state signal, constraint indicators, \\
missing-information signals, task structure, contextual signals from persona. \\

System-side axes: \\
product features, suitability signals, attribute alignment, price signals, \\
brand alignment, recommendation rationale, risk indicators. \\

Output format (JSON only): \\
\{ \\
\ \ "user\_aspects": [ \\
\ \ \ \ \{"name": str, "value": str, "justification": str\} \\
\ \ ], \\
\ \ "solution\_aspects": [ \\
\ \ \ \ \{"name": str, "value": str, "justification": str\} \\
\ \ ] \\
\} \\

Justifications must cite evidence from the input or be null if implicit.
}
\end{quote}

\section{Training Details for the Dimension Generating Agent}
\label{app:dga_training}

We fine-tune the Dimension Generating Agent (DGA) using the LLaMA-Factory framework~\citep{llamafactory2024}, which provides a unified interface for supervised fine-tuning of large language models with parameter-efficient adaptations. We experiment with two instruction-tuned backbone models: Meta-Llama-3-8B-Instruct~\citep{grattafiori2024llama} and Qwen-8B~\citep{qwen3}.

\paragraph{Fine-tuning Setup.}
The DGA is trained using supervised fine-tuning (SFT) with Low-Rank Adaptation (LoRA). We apply LoRA to all transformer layers with rank 8, allowing the model to adapt to dimension-level supervision while keeping the base model weights frozen. This design balances adaptation capacity and training efficiency.

\paragraph{Data and Input Formatting.}
Training is performed on the dimension-annotated datasets described in Appendix~A, using instruction-style prompts formatted according to the target model's native template (e.g., LLaMA-3). Inputs are truncated to a maximum context length of 3248 tokens. Data preprocessing is parallelized across 16 workers to support scalable training.

\paragraph{Optimization and Training Configuration.}
We train for 7 epochs using a cosine learning rate schedule with a warmup ratio of 0.1. The learning rate is set to $1\times10^{-4}$, with an effective batch size of 8 achieved via gradient accumulation. Training is conducted in bfloat16 precision. Models are evaluated every 500 steps on held-out validation splits to monitor learning dynamics.

\paragraph{Inference Use.}
At inference time, the fine-tuned DGA is prompted to infer task-relevant dimensions for unseen interactions. Importantly, implicit dimensions are not directly supervised during training but are elicited at inference by conditioning on the learned dimension vocabulary and patterns captured through explicit supervision.

\section{Prompt Specifications}
\label{app:prompts}

This section documents the prompt templates used across all components of our framework. We provide the full prompts for the Dimension Generating Agent (DGA) and the Response Generating Agent (RGA), including their domain-specific instantiations. These prompts encode the calibration principles described in the main paper and are listed here verbatim to support reproducibility and transparency.

\subsection{Dimension Generating Agent Prompts}
\label{app:dga_prompts}

Our DGA prompting strategy follows an expert-teacher analogy: a strong teacher does not merely echo what a student said, but uses prior experience with similar problems to identify what is \emph{structurally missing} in the student's formulation. Concretely, the prompts ask the model to (i) extract what the user has made explicit, and (ii) surface non-redundant, task-relevant dimensions that are commonly required for successful resolution of similar tasks but are not currently stated. To prevent overreach, the prompts explicitly prohibit restating the user's content and prohibit providing solutions or advice, ensuring the output is a clean dimension inventory rather than a response. Across domains, the prompts enforce a strict JSON interface, enabling direct use as supervision and as a controllable intermediate representation in downstream components.

We provide the exact prompts used for CodeContest, PersonalWAB, and Medical Diagnosis below.

\paragraph{Code Contests}
\begin{quote}
\small
\ttfamily{
You are an expert coding problem analyst. Your task is to analyze the user's coding query, understand the underlying goal, and identify both the dimensions they have already expressed and the important dimensions they have not yet mentioned.

Think carefully about how coding tasks with similar goals are typically specified and solved. More complete or successful formulations often depend on additional considerations that the current user has not made explicit. Your job is to surface these missing dimensions, not to solve the problem.

You must not repeat or restate any explicit information. You must not provide solutions, advice, code, or guidance. Focus only on identifying the conceptual structure of the problem.

Your output contains two types of dimensions:\\
---------------------------------------------------\\
explicit\_dimensions\\
---------------------------------------------------\\
These are problem-solving dimensions clearly grounded in the user query (e.g., stated goals, constraints, assumptions, or task structure).

Each explicit dimension must include:\\
  - name: concise conceptual label\\
  - value: information directly stated or implied in the query\\
  - justification: why this dimension is relevant to understanding or solving tasks of this type

Extract all explicit dimensions that meaningfully appear in the query.\\
--------------------------------------------------\\
missed\_dimensions\\
---------------------------------------------------\\
These are important considerations that the user has not mentioned but that typically matter when solving similar coding tasks. They should be conceptually diverse and non-redundant.

A missed dimension must:\\
- not appear in the query,\\
- not be safely inferable from explicit content,\\
- not be a generic personal detail (e.g., deadline, experience),\\
- represent a factor that affects correctness, robustness, efficiency, feasibility, resource use, or clarity.

Each missed dimension must include:\\
  - name: conceptual label (e.g., edge-case category, constraint type)\\
  - value: a plausible concrete instance or relevant detail\\
  - justification: why this dimension matters and how its absence leaves the problem incomplete 

You are encouraged to surface out-of-the-box dimensions that a typical user may never think to articulate. These should reveal deeper structural requirements, hidden dependencies, contextual contingencies, or evaluative axes that matter in realistic recommendation scenarios.\\
Extract as many missed dimensions as possible.  \\

---------------------------------------------------\\
OUTPUT FORMAT (STRICT)\\
---------------------------------------------------\\
Return ONLY:\\

===START\_JSON===\\
\{\\
  "explicit\_dimensions": [\\
    \{"name": "string", "value": "string", "justification": \\"string"\},\\
    ...\\
  ],\\
  "missed\_dimensions": [\\
    \{"name": "string", "value": "string", "justification": \\"string"\},\\
    ...\\
  ]\\
\}\\
===END\_JSON===\\

Do not output anything outside the JSON markers.\\

---------------------------------------------------\\
INPUT\\
---------------------------------------------------\\
User\_persona: <INSERT\_PERSONA>\\
User\_query: <INSERT\_USER\_QUERY>\\
Generate the JSON now.\\
}
\end{quote}

\paragraph{PersonalWAB}
\begin{quote}
\small
\ttfamily

You are an expert annotator analyzing user's query in personalized shopping and recommendation tasks. Your goal is to examine the user's query, infer their underlying intent, and identify both the dimensions they have explicitly expressed and the important dimensions they have not yet mentioned.

Think carefully about how shopping or preference-seeking tasks are typically formulated. More complete or successful recommendations often depend on additional considerations that the current user has not explicitly articulated. Your role is to surface these missing dimensions, not to solve the user's problem or recommend products.

You must not repeat or restate explicit information.  \\
You must not generate suggestions, advice, or recommendations.\\  
Focus only on identifying the conceptual structure of the user's query.

Your output MUST be valid JSON and nothing else.\\
You MUST start the JSON with: ===START\_JSON===\\
You MUST end the JSON with: ===END\_JSON===\\

No extra text, no explanations, no markdown, no trailing characters of any kind.\\

Your output contains two types of dimensions:\\
---------------------------------------------------\\
explicit\_dimensions\\
---------------------------------------------------\\
These are dimensions clearly grounded in the user query (such as stated needs, constraints, preferences, context, or user motivations).

Each explicit dimension must include:\\
  - name: concise conceptual label\\
  - value: information directly stated or implied in the query\\
  - justification: why this dimension is relevant for understanding shopping or preference-oriented tasks

Extract as many  explicit dimensions as possible that meaningfully appear in the query.\\
---------------------------------------------------\\
missed\_dimensions\\
---------------------------------------------------\\
These are important considerations not mentioned by the user but crucial for solving similar shopping or preference tasks. They should be conceptually diverse and non-redundant.

A missed dimension must:\\
- not appear in the query,\\
- not be safely inferable from the explicit content,\\
- not be a generic personal detail (e.g., experience level),\\
- represent a factor that affects relevance, usefulness, ranking,
  personalization, feasibility, constraints, or decision quality.

Each missed dimension must include:\\
  - name: conceptual label (e.g., budget constraints, lifecycle needs,product compatibility, preference specificity, situational factors)\\
  - value: a plausible concrete instance or detail\\
  - justification: why this dimension matters and how its absence leaves the preference or shopping need underspecified\\

Extract at least 10 meaningful missed dimensions.

---------------------------------------------------\\
OUTPUT FORMAT (STRICT)\\
---------------------------------------------------\\
Return ONLY:\\

===START\_JSON===\\
\{\\
  "explicit\_dimensions": [\\
    \{"name": "string", "value": "string", "justification":\\ "string"\}\\
  ],\\
  "missed\_dimensions": [\\
    \{"name": "string", "value": "string", "justification":\\
    "string"\}\\
  ]\\
\}\\
===END\_JSON===\\

Do not output anything outside the JSON markers.

---------------------------------------------------\\
INPUT\\
---------------------------------------------------\\
user\_query: \{user\_query\}\\

Generate the JSON now.\\
\end{quote}

\paragraph{Medical Diagnosis}
\begin{quote}
\small
\ttfamily
You are an expert clinical reasoning analyst. Your task is to study the patient\_query, understand the underlying health concern, and identify both the dimensions the patient has already expressed and the important clinical considerations they have not yet mentioned.

Think carefully about how clinicians typically interpret and structure similar patient presentations. More complete or successful formulations of medical concerns often depend on additional contextual, symptom-based, or risk-related factors that the patient has not stated. Your job is to surface these missing dimensions, not to diagnose or recommend treatment.

You must not repeat or restate any explicit information. You must not provide medical advice. Focus only on exposing the underlying
problem-structure: what is specified, and what remains unspecified. \\
---------------------------------------------------\\
1. explicit\_dimensions\\
---------------------------------------------------\\
These are aspects clearly grounded in the patient's text, such as:\\
- reported symptoms,\\
- subjective interpretations of severity,\\
- concerns or questions,\\
- contextual details (timeline, triggers),\\
- emotional or informational states.\\

Each explicit dimension must include:\\
  - name: concise clinical or contextual label\\
  - value: the value directly stated or clearly implied in the patient's text\\
  - justification: brief explanation of why this dimension is relevant for understanding similar medical presentations\\

Extract all meaningful explicit dimensions. Do not invent details.

---------------------------------------------------\\
2. missed\_dimensions\\
---------------------------------------------------\\
These are important clinical considerations the patient has not mentioned but that clinicians typically explore when evaluating similar symptoms.\\
They should be conceptually diverse and non-redundant.\\

A missed dimension must:\\
- NOT appear in the patient's text,\\
- NOT be safely inferable,\\
- NOT be a generic personal detail unrelated to the complaint,\\
- reflect a clinically relevant factor affecting interpretation of symptoms,\\
  risk assessment, differential considerations, or need for follow-up.\\

Useful categories often include:
- missing symptom qualifiers (duration, progression, distribution),\\
- unreported associated symptoms or red flags,\\
- gaps in exposure or risk factors,\\
- missing information about onset, triggers, or patterns,
- relevant medical history elements,\\
- unasked clarifying questions clinicians typically use to assess severity.\\

Each missed dimension must include:\\
  - name: concise conceptual label (e.g., "Associated Symptoms", "Exposure History")\\
  - value: a plausible clinically relevant instance (e.g., “fever”, “recent contact”,“sudden vs gradual onset”, “immune status”)\\
  - justification: why this dimension and value matter for interpreting the presentation and how their absence limits understanding\\

Produce a diverse, non-overlapping set of missed dimensions.

---------------------------------------------------\\
OUTPUT FORMAT (STRICT)\\
---------------------------------------------------\\
Return ONLY:\\

===START\_JSON===\\
\{\\
  "explicit\_dimensions": [\\
    \{"name": "string", "value": "string", "justification":\\ "string"\},\\
    ...\\
  ],\\
  "missed\_dimensions": [\\
    \{"name": "string", "value": "string", "justification": \\"string"\},\\
    ...\\
  ]\\
\}\\
===END\_JSON===\\

Do not output anything outside the JSON markers.\\
---------------------------------------------------\\
INPUT\\
---------------------------------------------------\\
patient\_query: \{patient\_query\}\\

Generate the JSON now.\\
\end{quote}

\subsection{Response Generating Agent Prompts}
\label{app:rga_prompts}

While the DGA surfaces \emph{what is missing}, the Response Generating Agent (RGA) governs \emph{how and when} those missing dimensions should be addressed. We design RGA prompts around an expert-editor analogy: an expert does not rewrite an answer from scratch, but carefully revises an existing response by filling gaps, correcting omissions, and calibrating tone and scope to the user's intent. The prompts therefore frame response generation as a \emph{selective update} to a baseline response, explicitly conditioning on (i) the original user query, (ii) the existing system response, and (iii) the activated dimensions identified by the reranker.

Across domains, the prompts enforce three core principles. First, \textbf{initiative calibration}: the model must infer how proactive to be from the query itself rather than uniformly expanding the response. Second, \textbf{constraint-preserving repair}: the original response's structure and intent are preserved unless clarity, safety, or personalization requires otherwise. Third, \textbf{controlled uncertainty handling}: implicit dimensions are incorporated only when appropriate, with at most one clarifying question allowed to avoid unwarranted assumptions. Together, these constraints operationalize proactivity as a controlled, context-sensitive decision rather than verbosity. We provide the exact RGA prompts used for CodeContest, Medical Diagnosis, and PersonalWAB below.

\paragraph{Code Contest.}
\begin{quote}
\small
\ttfamily

You are acting as a proactive coding assistant that reflects on an earlier response from a coding assistant.\\
\\
Task:\\
Given a user query, an existing coding assistant response, and a set of missing aspects, generate an updated response that better supports the user by addressing relevant uncertainty and unmet informational needs.\\
\\
Guiding principle:\\
The appropriate level of guidance depends on the patient's question.\\
You must infer how proactive to be from the query itself.\\
\\
Initiative calibration:\\
If the query suggests confusion, failure, or blocked progress, expand the response to proactively cover important missing aspects.\\
If the query is narrow or explicitly limited, keep the response focused and minimally expanded.\\
If the query is narrow or clearly constrained, keep the response focused and minimally expanded.\\
If the query reflects frustration or ambiguity, broaden gently while avoiding assumptions about user skill or intent.\\
\\
Handling missing aspects:\\
Explicit missing aspects should generally be addressed.\\
Implicit missing aspects represent possible knowledge gaps and should be included only when appropriate for this query.\\
If addressing an implicit aspect requires user-specific details, ask at most ONE clarifying question rather than speculating.\\
\\
Response constraints:\\
Maintain a supportive, non-authoritative tone.\\
Avoid condescension, over-specification, or unnecessary implementation detail.\\
Preserve the original response's structure and intent unless safety or clarity requires change.\\
Prefer concise additions over complete rewrites.\\
\\
Output constraints:\\
You MUST start with: ===START===\\
You MUST end with: ===END===\\
Output ONLY the final clinical assistant response.\\
No explanations,\\
no markdown, no trailing characters.\\
\\
User query:\\
<user\_query>\\
\\
Existing coding assistant response:\\
<system\_output>\\
\\
Missing aspects (explicit missing + implicit knowledge gaps):\\
<missed\_aspects>
\end{quote}

\paragraph{Medical Diagnosis}
\begin{quote}
\small
\ttfamily
You are acting as a clinical assistant that reflects on an existing response to a patient.\\
\\
Task:\\
Given a patient query, an existing clinical assistant response, and a set of missing aspects, generate an updated response that better supports the patient by addressing relevant uncertainty and unmet informational needs.\\
\\
Guiding principle:\\
The appropriate level of guidance depends on the patient's question.\\
You must infer how proactive to be from the query itself.\\
\\
Initiative calibration:\\
If the query suggests urgency, safety risk, or possible harm, expand the response to proactively cover important missing aspects.\\
If the query is narrow or explicitly limited, keep the response focused and minimally expanded.\\
If the query signals uncertainty, curiosity, or lack of clarity, selectively surface helpful missing context.\\
If the query signals emotional strain or ambiguity, broaden gently while avoiding assumptions or diagnoses.\\
\\
Handling missing aspects:\\
Explicit missing aspects should generally be addressed.\\
Implicit missing aspects represent possible knowledge gaps and should be included only when appropriate for this query.\\
If addressing an implicit aspect requires patient-specific details, ask at most ONE clarifying question rather than speculating.\\
\\
Response constraints:\\
Maintain a supportive, non-authoritative clinical tone.\\
Avoid diagnosis, definitive medical claims, or alarming language.\\
Preserve the original response's structure and intent unless safety or clarity requires change.\\
Prefer concise additions over complete rewritesnd complete rewrites.\\
\\
Output constraints:\\
You MUST start with: ===START===\\
You MUST end with: ===END===\\
Output ONLY the final clinical assistant response.\\
No explanations,\\
no markdown, no trailing characters.\\
\\
Patient query:\\
<user\_query>\\
Existing clinical assistant response:\\
<system\_output>\\
Missing aspects (explicit missing + implicit knowledge gaps):\\
<missed\_aspects>
\end{quote}

\paragraph{PersonalWAB}
\begin{quote}
\small
\ttfamily
You are acting as a proactive recommendation assistant that reflects on an earlier response from a recommendation assistant.\\
\\
Task:\\
Given a user query, an existing recommendation assistant response, and a set of missing aspects, generate an updated response that better supports the user by addressing relevant uncertainty and unmet informational needs.\\
\\
Guiding principle:\\
The appropriate level of guidance depends on the user's question.\\
You must infer how proactive to be from the query itself.\\
\\
Initiative calibration:\\
If the query suggests indecision, dissatisfaction, or conflicting preferences, expand the response to proactively cover important missing aspects.\\
If the query is clearly scoped or asks for a specific type of suggestion, keep the response focused and minimally expanded.\\
If the query signals curiosity, vague goals, or exploratory intent, selectively surface relevant tradeoffs or options.\\
If the query reflects overwhelm or uncertainty, broaden gently while avoiding assumptions about user preferences or constraints.\\
\\
Handling missing aspects:\\
Explicit missing aspects should generally be addressed.\\
Implicit missing aspects represent possible preference gaps or context needs and should be included only when appropriate for this query.\\
If addressing an implicit aspect requires user-specific details, ask at most ONE clarifying question rather than speculating.\\
\\
Response constraints:\\
Maintain a helpful, user-centered tone.\\
Avoid prescriptive, one-size-fits-all suggestions or rigid criteria.\\
Preserve the original response's structure and intent unless personalization or clarity requires change.\\
Prefer concise additions over complete rewrites.\\
\\
Output constraints:\\
You MUST start with: ===START===\\
You MUST end with: ===END===\\
Output ONLY the final clinical assistant response.\\
No explanations,\\
no markdown, no trailing characters.\\
\\
User query:\\
<user\_query>\\
\\
Existing recommendation assistant response:\\
<system\_output>\\
\\
Missing aspects (explicit missing + implicit knowledge gaps):\\
<missed\_aspects>
\end{quote}

\subsection{Evaluation Prompt}
\label{app:evaluation_prompt}

Evaluating calibrated proactivity requires reasoning beyond surface-level correctness or fluency. The evaluator must infer user intent, identify implicit risks or missing considerations, and judge whether the assistant exercised the \emph{appropriate amount of initiative} for the inferred domain. To this end, we use GPT-5 as an evaluation model due to its strong performance in instruction following, domain inference, and nuanced judgment across heterogeneous tasks. Importantly, GPT-5 is not used to generate responses in our framework; rather, it serves as a consistent and high-capacity evaluator to assess response quality under a unified rubric.

The evaluation prompt is designed to avoid common pitfalls in pairwise comparison. Responses are evaluated \emph{independently}, without assuming which system is baseline or proactive, and without directly comparing responses against each other. This design prevents preference leakage and encourages absolute judgments grounded in the user's needs rather than relative differences. Proactivity is explicitly framed as \emph{calibration}, not verbosity: evaluators are instructed to penalize both overreach (unwarranted assumptions, excessive warnings) and underreach (failure to address salient risks or gaps).

Each response is scored holistically on a 0--5 scale, considering intent alignment, anticipation of implicit needs, appropriateness of initiative, personalization, clarity, and missed opportunities. The requirement for concise, single-line justifications enforces disciplined reasoning and reduces evaluator drift, while the strict JSON output format ensures reliable downstream aggregation and analysis.

\paragraph{Evaluation Prompt}
\begin{quote}
\small
\ttfamily
You are an expert evaluator of proactive and personalized AI assistants.\\
Your task is to independently evaluate two assistant responses to a user query with respect to calibrated proactivity and personalization.\\
You will be given:\\
- A user query\\
- Response A\\
- Response B\\
\\
You must infer:\\
- The interaction domain (e.g., medical, shopping, coding, or other)\\
- The user's explicit needs\\
- Any important implicit considerations (e.g., risks, uncertainties, missing context, next steps)\\
\\
Evaluation Principles:\\
- Evaluate each response independently. Do NOT compare them directly.\\
- Do NOT assume which response is baseline or proactive.\\
- Proactivity is not verbosity; it is taking the right amount of initiative given the user's intent and inferred domain.\\
- Penalize overreach (irrelevant warnings, excessive assumptions, unnecessary steps).\\
- Penalize underreach (ignoring important needs, risks, or uncertainties).\\
\\
Score each response holistically on a 0--5 scale considering:\\
1. How well it addresses the user's stated intent\\
2. How well it anticipates and handles unspoken but relevant needs\\
3. Appropriateness of initiative for the inferred domain\\
4. Personalization and contextual relevance\\
5. Clarity, tone, and overall usefulness\\
6. Missed opportunities or unnecessary intervention\\
\\
Scoring Scale:\\
0 -- Very poor\\
1 -- Poor\\
2 -- Weak\\
3 -- Moderate\\
4 -- Good\\
5 -- Excellent\\
\\
Output Requirements:\\
- Output MUST be valid JSON ONLY.\\
- Do NOT include explanations outside the JSON.\\
- Do NOT add extra fields.\\
- Justifications must be a single concise line each.\\
\\
Respond ONLY in the following format:\\
\\
\{\\
\ \ "response\_A\_score": <integer 0--5>,\\
\ \ "response\_B\_score": <integer 0--5>,\\
\ \ "response\_A\_justification": "<single concise line>",\\
\ \ "response\_B\_justification": "<single concise line>"\\
\}\\
\\
User Query:\\
<INSERT USER QUERY HERE>\\
Response A:\\
<INSERT RESPONSE A HERE>\\
Response B:\\
<INSERT RESPONSE B HERE>
\end{quote}

\subsection{Chain of Thought Prompts}
\label{app:cot_prompt}
We employ a two-stage chain-of-thought (CoT) pipeline. In the first stage, the model analyzes the user query and an initial response to identify explicit and implicit informational aspects that were missed or under-addressed. In the second stage, these aspects are used as structured guidance to generate a refined response that better aligns with the user's intent, uncertainty, and safety requirements.

\xhdr{Prompt 1} This prompt is used to analyze a medical user query and an initial model response in order to identify explicit and implicit informational aspects that were missed or insufficiently addressed. The extracted aspects serve as structured inputs for downstream response refinement.
\begin{quote}
\small
\ttfamily
You are an expert annotator analyzing clinical question--answer interactions.\\
\\
Task:\\
Given a medical user query and an existing clinical assistant response,\\
extract medically relevant aspects that were:\\
(a) explicitly requested but not adequately addressed, and/or\\
(b) implicitly relevant but reasonably missing given the query.\\
\\
Rules:\\
-- Do NOT invent facts.\\
-- Infer latent aspects only when weakly but text-grounded.\\
-- Do NOT judge style or tone---focus only on medical/technical content.\\
-- Output structured aspects only.\\
\\
Inputs\\
query: <INSERT\_USER\_QUERY>\\
solution: <INSERT\_EXISTING\_RESPONSE>\\
\\
Aspect definition\\
aspect = \{\\
\ \ "name": string,\\
\ \ "value": string,\\
\ \ "justification": string\\
\}\\
\\
-- justification must quote a verbatim substring from the input,\\
\ \ or ``null'' if the aspect is implicit.\\
\\
Canonical aspect types (preferred, but not limited to)\\
\\
User aspects:\\
-- clinical\_goal\_intent\\
-- explicit\_symptom\\
-- explicit\_history\\
-- latent\_symptom\_pattern\\
-- disease\_specific\_indicator\\
-- multisystem\_interaction\\
-- risk\_indicator\\
-- constraint\_indicator\\
-- missing\_information\\
-- treatment\_history\_signal\\
-- emotional\_state\\
-- task\_structure\\
\\
System (response) aspects:\\
-- medical\_reasoning\_pattern\\
-- disease\_specific\_assessment\\
-- explicit\_guidance\\
-- treatment\_guidance\\
-- safety\_statement\\
-- diagnostic\_hypothesis\\
-- risk\_assessment\\
-- uncertainty\_handling\\
-- reassurance\_strategy\\
-- information\_gap\_response\\
-- guideline\_alignment\\
-- correctness\_risk\\
\\
Output schema (JSON ONLY)\\
\{\\
\ \ "user\_aspects": [\\
\ \ \ \ \{"name": str, "value": str, "justification": str\}\\
\ \ ],\\
\ \ "solution\_aspects": [\\
\ \ \ \ \{"name": str, "value": str, "justification": str\}\\
\ \ ]\\
\}\\
\\
Constraints:\\
-- Produce as many distinct, medically meaningful aspects as supported.\\
-- Remove duplicates.\\
-- No markdown, no explanation, JSON only.
\end{quote}

\xhdr{Prompt 2} This prompt is used to generate an improved clinical response by incorporating the previously identified missing aspects, while calibrating initiative and medical caution to the intent and risk level expressed in the original user query.

\begin{quote}
\small
\ttfamily
You are acting as a clinical assistant reflecting on an existing response\\
to a patient.\\
\\
Task:\\
Given:\\
1) a patient query,\\
2) an existing clinical assistant response,\\
3) a list of missing aspects (explicit + implicit),\\
\\
generate an updated response that better supports the patient by addressing\\
relevant uncertainty and unmet informational needs.\\
\\
Guiding principle:\\
Calibrate initiative based on the patient's query.\\
\\
Initiative calibration:\\
-- If the query suggests urgency, safety risk, or possible harm,\\
\ \ proactively address critical missing aspects.\\
-- If the query is narrow or explicitly limited,\\
\ \ keep the response focused and minimally expanded.\\
-- If the query signals uncertainty or curiosity,\\
\ \ selectively surface helpful missing context.\\
-- If the query signals emotional strain or ambiguity,\\
\ \ broaden gently without assumptions or diagnoses.\\
\\
Handling missing aspects:\\
-- Explicit missing aspects $\rightarrow$ generally address.\\
-- Implicit missing aspects $\rightarrow$ include only if appropriate.\\
-- If an implicit aspect requires patient-specific info,\\
\ \ ask AT MOST ONE clarifying question.\\
\\
Response constraints:\\
-- Supportive, non-authoritative clinical tone.\\
-- No diagnosis, no definitive medical claims.\\
-- Avoid alarming language.\\
-- Preserve original structure and intent unless safety requires change.\\
-- Prefer concise additions over full rewrites.\\
\\
Output constraints:\\
-- MUST start with: ===START===\\
-- MUST end with: ===END===\\
-- Output ONLY the final clinical assistant response.\\
-- No explanations, no markdown, no extra text.\\
\\
Inputs:\\
Patient query:\\
<user\_query>\\
\\
Existing clinical assistant response:\\
<system\_output>\\
\\
Missing aspects:\\
<missed\_aspects>
\end{quote}

\label{app:additional_experiments}
\end{document}